\documentclass{article}

\usepackage[final]{corl_2026} 

\usepackage[dvipsnames,table]{xcolor}
\usepackage{graphicx}
\usepackage{amsmath}
\usepackage{amssymb}
\usepackage{nicefrac}
\usepackage{mathtools}
\usepackage{hyperref}
\usepackage{enumitem}
\usepackage{xfrac}
\usepackage{siunitx}
\usepackage{graphicx}
\usepackage{subcaption}
\usepackage{wrapfig}
\usepackage{booktabs}
\usepackage{multirow}
\usepackage{makecell}
\usepackage{array}
\usepackage{adjustbox}
\usepackage{pifont}

\newcommand{\solidrightarrow}{\relbar\joinrel\blacktriangleright}

\hypersetup{
    colorlinks=true,
    linkcolor=brown,
    citecolor=RoyalPurple!75,
    urlcolor=Maroon,
    pdfauthor={Youwei Yu, Lantao Liu},
    pdftitle={Generative Flow on Symbolic Abstractions for Embodiment Capability Discovery}
}

\title{Generative Flow on Symbolic Abstractions for\\Embodiment Capability Discovery}

\author{
  Youwei Yu \quad Lantao Liu\\
 Indiana University Bloomington\\
  \texttt{\{youwyu, lantao\}@iu.edu}
}

\begin{document}
\maketitle


\begin{figure}[h!]
    \vspace*{-.3in}
    \centering
    \includegraphics[width=.85\linewidth]{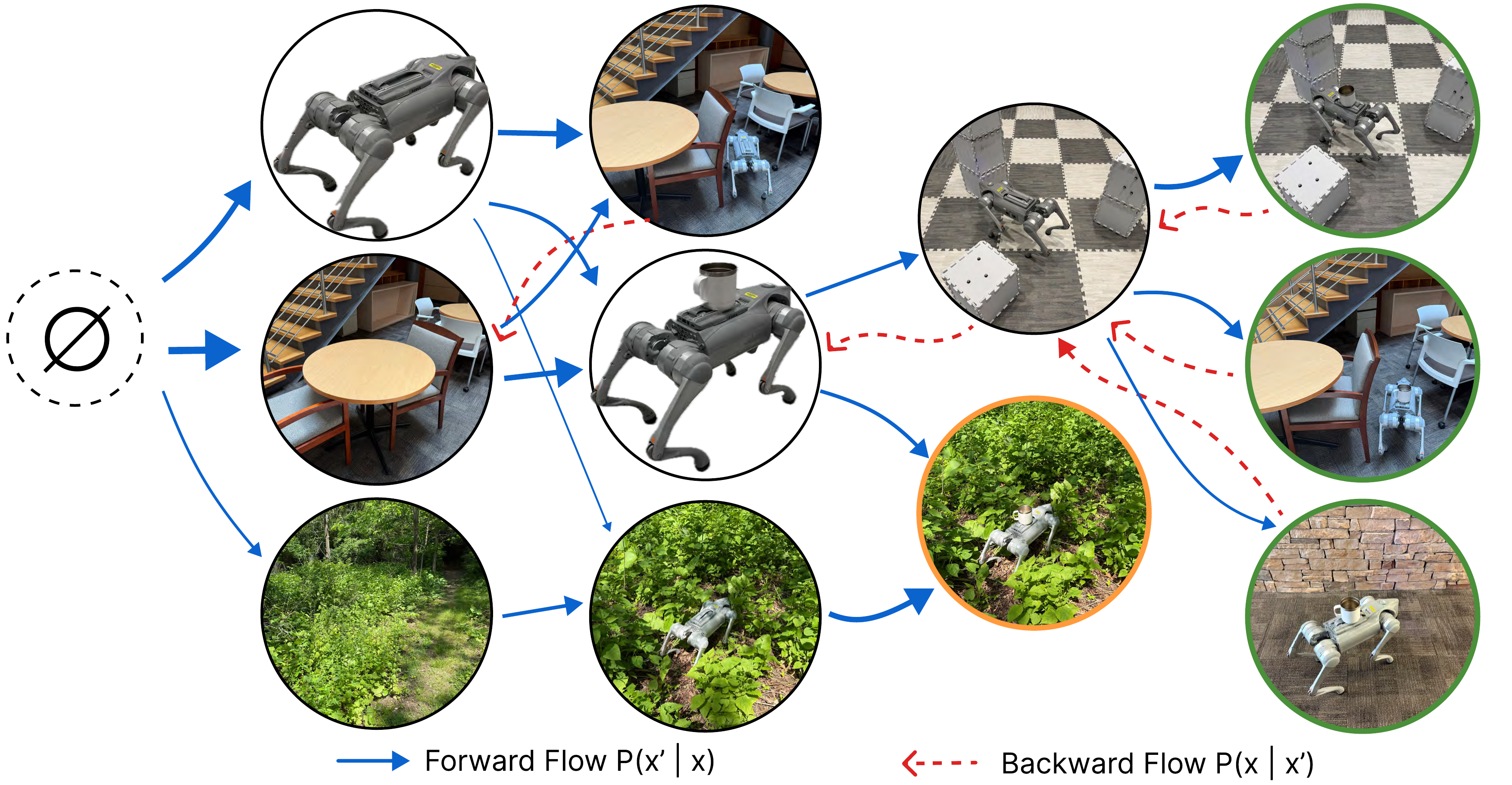}
    \vspace*{-2.5mm}
    \caption{\small
    FOSA learns a generative flow policy over symbolic task skeletons to discover tasks near a robot's empirical capability frontier. Starting from the empty skeleton $\varnothing$, the policy samples grammar-valid semantic edits with transition probability $P(\mathbf{x}'\mid\mathbf{x})$. Blue arrow $\textcolor{blue}{\solidrightarrow}$ thickness indicates sampling probability.
    }
    \label{fig:teaser}
    \vspace*{-1mm}
\end{figure}


\begin{abstract}
Achieving complex, long-horizon robot behavior remains challenging because successful execution depends on skills grounded in a robot's physical affordances. Existing training-task distributions are often handcrafted or sampled from large language models (LLMs), relying on semantic plausibility rather than embodiment-specific feasibility. This mismatch can produce curricula that are too difficult, redundant, or blind to non-obvious but learnable behaviors. We propose Generative Flow on Symbolic Abstractions (FOSA), a framework for discovering embodiment-specific capabilities by learning a grounded task generator. FOSA parameterizes an abstract policy as a Generative Flow Network that constructs symbolic task skeletons. Each skeleton is compiled into a simulation environment, where low-level policy-learning performance provides the reward signal for the abstract policy. Across heterogeneous robot platforms, FOSA discovers diverse and executable tasks near the robot's empirical capability frontier. After distilling the discovered task policies into a single language-conditioned policy, FOSA improves performance on unseen complex and long-horizon tasks. The code and demonstration videos are available at \url{https://www.youwei-yu.com/fosa}.
\end{abstract}



\section{Introduction}
\label{sec:intro}

Robot learning has become increasingly effective once the target task is specified and lies within the robot's learnable regime. With a well-specified reward and simulator, reinforcement learning (RL) with domain and dynamics randomization can produce agile policies that transfer from simulation to hardware~\cite{tobin2017domain,peng2018sim,tan2018sim,hwangbo2019learning,kumar2021rma}. With demonstrations, expressive policy classes can represent multimodal behaviors in contact-rich manipulation~\cite{chi2025diffusion}. These successes still presuppose that the task distribution has been chosen. In practice, robot embodiments are usually designed for broad classes of use rather than a single task, leaving their actual capability space only partially explored.

The choice of task distribution becomes critical for complex and long-horizon policy learning. Skill composition, hierarchical RL, and task-and-motion planning all rely on intermediate capabilities that can be invoked, chained, or refined~\cite{sutton1999between,bacon2017option,garrett2021integrated,li2025bilevel}. Yet these capabilities are often assumed known, manually designed, or proposed from human semantic intuition. These human-specified structures are brittle across embodiments: a task that appears simple may be physically difficult for a robot, while a non-obvious behavior may be easy to learn and useful for composition~\cite{huang2026kinder}.

We study \emph{embodiment capability discovery}: finding the task structures that a specific robot can learn under a fixed training budget. The central object is the robot's empirical capability frontier, consisting of tasks that are learnable but not already trivial. This frontier is useful for two reasons. First, frontier tasks expose skills that can support interpolation, composition, and downstream long-horizon generalization. Second, the full task space is combinatorial, so exhaustive enumeration is infeasible; efficient discovery requires learning where useful capabilities are likely to emerge.

We propose Generative \textbf{F}low \textbf{o}n \textbf{S}ymbolic \textbf{A}bstractions (FOSA), a framework that learns an abstract task generator over symbolic task skeletons. FOSA samples a skeleton, compiles it into a simulation environment, trains a low-level task policy, and uses the resulting performance to update the generator. We instantiate the generator as a Generative Flow Network~\cite{bengio2023gflownet}, enabling diverse sampling over variable-length and variable-dimensional task specifications near the capability frontier. Language models are used as compilation assistants for converting symbolic skeletons into executable environments.
Our contributions are summarized as follows:
\begin{itemize}[leftmargin=*] \vspace*{-5pt}
    \item We propose FOSA, a general framework for embodiment capability discovery that learns an abstract policy over symbolic task skeletons and grounds task selection in low-level policy-learning performance.

    \item We instantiate the abstract policy with a Generative Flow Network, enabling diverse search over variable-length and variable-dimensional task specifications near the robot's empirical capability frontier.

    \item We evaluate FOSA across heterogeneous robot platforms in simulation and the real world. We find that proximity to the discovered capability frontier correlates with generalization to held-out complex and long-horizon tasks, and that FOSA improves performance compared with task-generation and curriculum-learning baselines.
\end{itemize}


\section{Related Work}
\label{sec:literature}

\textbf{Complex and Long-Horizon Robot Behavior.}
Complex robot policy often requires temporal abstraction, reusable skills, and composition over extended horizons. Hierarchical RL and options provide classical mechanisms for learning and invoking temporally extended actions~\cite{sutton1999between,bacon2017option,nachum2018data}, while skill-prior and skill-chaining methods improve long-horizon control by reusing learned behavioral primitives~\cite{gupta2019relay,pertsch2020spirl,ajay2021opal,lee2021adversarial}. Task-and-motion planning and bilevel formulations similarly decompose long-horizon problems into symbolic decisions and continuous control problems~\cite{garrett2021integrated,li2025bilevel}. In robotics, platform-specific learning has produced increasingly rich behaviors, including agile quadruped locomotion, dexterous manipulation, and articulated-object interaction~\cite{hwangbo2019learning,akkaya2019solving,cheng2023extreme,hoeller2023learning,bao2023dexart}. These works demonstrate the importance of skills and compositional structure, while typically assuming that the relevant task family, skill library, or decomposition is already specified.

\textbf{Skill Discovery and Open-Ended Curricula.}
A complementary line of work studies how agents can acquire skills, goals, or curricula with less manual supervision. Unsupervised skill discovery learns diverse behaviors through intrinsic objectives based on mutual information, state coverage, empowerment, contrastive representation learning, or predictable skill-conditioned dynamics~\cite{eysenbach2018diversity,campos2020explore,laskin2022cic,liu2021aps}. Goal-generation and curriculum-learning methods adapt the training distribution according to learning progress, regret, success rate, or task difficulty~\cite{florensa2018automatic,portelas2020automatic,zhang2020automatic}. Open-ended environment design further studies how to generate environments that induce increasingly complex agent behavior~\cite{wang2019poet,dennis2020emergent,jiang2021prioritized,parker2022evolving,tio2023transferable}. These methods reduce manual ordering of experience, but often operate within a fixed task family, fixed simulator parameterization, or motor-level behavior space. FOSA instead searches over symbolic task skeletons and scores each candidate through embodiment-specific policy learning.

\textbf{Language-Driven Generation and Scalable Robot Learning.}
Large language models have recently been used to scale robot learning by generating tasks, rewards, curricula, code, simulation scenes, and supervision~\cite{huang2022inner,liang2023code,wang2023gen,wang2023robogen,ma2023eureka,xie2023text2reward,ma2024dreureka,liang2024eurekaverse,xia2026sage}. These systems reduce manual engineering and make it possible to instantiate more diverse robot-learning problems. In parallel, sim-to-real methods improve transfer through domain randomization, dynamics randomization, system identification, and online adaptation~\cite{tobin2017domain,peng2018sim,tan2018sim,hwangbo2019learning,kumar2021rma}. Recent robot foundation models and vision-language-action policies scale policy learning with large heterogeneous robot datasets~\cite{brohan2022rt1,brohan2023rt2,kim2025openvla,kim2025fine}, while morphology-conditioned control and robot-design methods show that policy learning is strongly coupled to embodiment~\cite{zhao2020robogrammar,yuan2021transform2act,gupta2022metamorph}.
FOSA uses the LLM only to compile skeletons and propose edits; task retention is decided by embodiment-grounded policy learning.


\section{Preliminaries}
\label{sec:preliminaries}

We formulate robot capability discovery as learning an abstract policy over symbolic task skeletons and grounding each sampled skeleton through low-level policy learning. We first define the task abstraction used by FOSA, then review the GFlowNet objective used to sample diverse skeletons in proportion to embodiment-grounded utility.

\textbf{Symbolic Abstractions.}
We represent each candidate robot-learning problem as a symbolic task skeleton, following the spirit of task-and-motion planning (TAMP)~\cite{garrett2021integrated,li2025bilevel}. The symbolic level specifies objects, goals, and operators, while the continuous level specifies the geometric, dynamical, and simulator parameters required for execution.
Let $\mathcal{Y}$ denote entity types, $\mathcal{P}$ predicate schemas, $\mathcal{U}$ operator schemas, and $\Phi$ feasibility constraints. For a finite set of typed entities $E$, let $\mathrm{Atoms}(\mathcal{P},E)$ and $\mathrm{Lits}(\mathcal{P},E)$ denote the induced ground atoms and literals. A task skeleton is defined as
\begin{equation}
\label{eq:skeleton}
    \mathbf{x}=[E_{\mathbf{x}},I_{\mathbf{x}},G_{\mathbf{x}},U_{\mathbf{x}},\boldsymbol{\theta}_{\mathbf{x}},\Phi_{\mathbf{x}}],
\end{equation}
where $E_{\mathbf{x}}$ is the set of typed task entities, $I_{\mathbf{x}}\subseteq \mathrm{Atoms}(\mathcal{P},E_{\mathbf{x}})$ is the initial symbolic state, $G_{\mathbf{x}}\subseteq \mathrm{Lits}(\mathcal{P},E_{\mathbf{x}})$ is the goal condition, $U_{\mathbf{x}}\subseteq\mathcal{U}$ is the set of referenced operator schemas, $\boldsymbol{\theta}_{\mathbf{x}}$ stores continuous arguments, and $\Phi_{\mathbf{x}}\subseteq\Phi$ stores feasibility constraints such as valid poses, contact modes, tolerances, and parameter ranges. For example, in a quadruped slippery-slope locomotion task, $E_{\mathbf{x}}$ contains the robot and ramp, $G_{\mathbf{x}}$ encodes velocity tracking, $\boldsymbol{\theta}_{\mathbf{x}}$ stores the incline and friction coefficient, and $\Phi_{\mathbf{x}}$ constrains feasible contacts and termination thresholds.

Given a robot embodiment $\mathbf{e}$, a complete and valid skeleton is compiled into an executable RL environment
\begin{equation}
\label{eq:env-compile}
    \mathcal{M}_{\mathbf{e},\mathbf{x}}
    =\mathrm{Env}(\mathbf{e},\mathbf{x})
    =(\mathcal{O}_{\mathbf{e},\mathbf{x}},\mathcal{A}_{\mathbf{e},\mathbf{x}},p_{\mathbf{e},\mathbf{x}},r_{\mathbf{e},\mathbf{x}},\rho_{\mathbf{e},\mathbf{x}},\Gamma_{\mathbf{e},\mathbf{x}}),
\end{equation}
where the tuple denotes observation space, action space, transition dynamics, reward, reset distribution, and termination conditions. A skeleton is valid if it is type-consistent and supported by the embodiment and simulator. It is complete if all information needed for policy training is specified.

\textbf{Capability Frontier.}
For a completed skeleton $\mathbf{x}$, FOSA trains a task policy $\boldsymbol{\pi}_{\mathbf{x}}$ in $\mathcal{M}_{\mathbf{e},\mathbf{x}}$ under a fixed sample budget and records its success rate before training, $\mathcal{J}^{0}_{\mathbf{x}}$, and after training, $\mathcal{J}_{\mathbf{x}}$, both in $[0,1]$. We define the empirical capability frontier as the tasks whose success rate training improves but does not saturate,
\begin{equation}
\label{eq:frontier-set}
    \mathcal{B}_{\mathbf{e}}
    =\{\mathbf{x}: \mathcal{J}^{0}_{\mathbf{x}}<\mathcal{J}_{\mathbf{x}}<1\}.
\end{equation}
Tasks on this frontier are learnable but nontrivial, and therefore provide informative feedback for capability discovery.

\textbf{Generative Flow Networks.}
A Generative Flow Network (GFlowNet)~\cite{bengio2023gflownet} learns a stochastic construction policy whose terminal samples are distributed in proportion to a positive reward function. In FOSA, the construction graph is a directed acyclic graph over partial skeletons. A trajectory
\begin{equation}
    \boldsymbol{\tau}=(\mathbf{x}^{0}\rightarrow\mathbf{x}^{1}\rightarrow\cdots\rightarrow\mathbf{x}^{T}=\mathbf{x}),
    \qquad \mathbf{x}^{0}=\varnothing,
\end{equation}
sequentially edits a partial skeleton until it reaches a complete skeleton. The forward policy $P_F(\mathbf{x}^{t+1}\mid\mathbf{x}^{t})$ samples the next valid edit, and the backward policy $P_B(\mathbf{x}^{t}\mid\mathbf{x}^{t+1})$ scores valid predecessor states. Given terminal reward $R(\mathbf{x})>0$, the target distribution is $P^{\star}(\mathbf{x})\propto R(\mathbf{x})$. FOSA trains the abstract policy using trajectory balance~\cite{malkin2022trajectory}:
\begin{equation}
\label{eq:tb-loss}
    \mathcal{L}_{\mathrm{GFlowNet}}(\boldsymbol{\tau})
    =
    \left(
    \log Z+\log P_F(\boldsymbol{\tau})
    -\log R(\mathbf{x})
    -\log P_B(\boldsymbol{\tau})
    \right)^2,
\end{equation}
where $P_F(\boldsymbol{\tau})=\prod_{t=0}^{T-1}P_F(\mathbf{x}^{t+1}\mid\mathbf{x}^{t})$, $P_B(\boldsymbol{\tau})=\prod_{t=0}^{T-1}P_B(\mathbf{x}^{t}\mid\mathbf{x}^{t+1})$, and $Z$ is a learned scalar.


\section{Method}
\label{sec:method}

FOSA discovers embodiment-specific capabilities by coupling an abstract policy over task skeletons with task policies optimized in generated environments. Given a robot embodiment $\mathbf{e}$, a symbolic domain $(\mathcal{Y},\mathcal{P},\mathcal{U},\Phi)$, a compiler $\mathrm{Env}(\cdot)$, and a fixed evaluation budget, FOSA returns an archive of discovered skeletons and trained policies. The archive is subsequently distilled into a general language-conditioned policy.

\begin{figure}[h!]
    \vspace*{-0.15in}
    \centering
    \includegraphics[width=.95\linewidth]{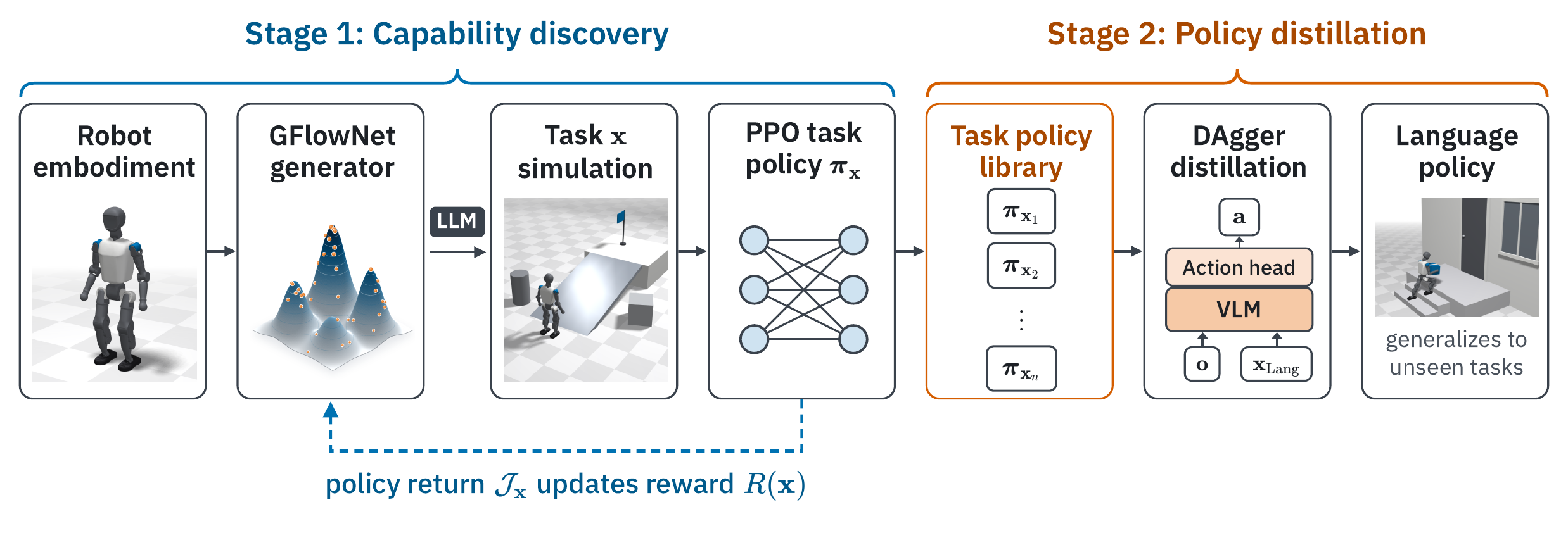}
    \vspace*{-0.15in}
    \caption{\small
    FOSA framework overview. The GFlowNet abstract policy generates symbolic task skeletons, which are compiled by an LLM into simulation environments to train task policies. Task-policy performance provides the reward signal that updates the abstract policy toward the robot's empirical capability frontier.
    }
    \label{fig:method}
    \vspace*{-4mm}
\end{figure}

\subsection{FOSA Framework}
\label{sec:fosa_system}

At each discovery iteration, the abstract policy samples a construction trajectory from $\mathbf{x}^{0}=\varnothing$ to a terminal skeleton $\mathbf{x}$. Deterministic validity and completeness checks reject skeletons that violate grammar, simulator, or embodiment constraints. The compiler instantiates $\mathcal{M}_{\mathbf{e},\mathbf{x}}=\mathrm{Env}(\mathbf{e},\mathbf{x})$. A task policy $\boldsymbol{\pi}_{\mathbf{x}}$ is trained from scratch in this environment using PPO~\cite{schulman2017proximal}; the success rate of the first evaluation is $\mathcal{J}^{0}_{\mathbf{x}}$ and that of the last is $\mathcal{J}_{\mathbf{x}}$. FOSA converts them into $R(\mathbf{x})$, updates the GFlowNet via Eq.~\eqref{eq:tb-loss}, and adds non-duplicate successful skeletons to the archive.

The LLM enters FOSA in two roles, neither of which selects tasks. Inside $\mathrm{Env}(\cdot)$ it is a compiler: from the verbalized and serialized skeleton it writes a staged JSON specification (e.g., simulation contract and reward)~\cite{wang2023gen,wang2023robogen,ma2024dreureka}. During sampling, it also supplies a prior over valid edits. Which skeletons are retained and reinforced is decided solely by $R(\mathbf{x})$, so FOSA is grounded by physical policy-learning performance rather than by semantic plausibility.

\subsection{FOSA Abstract Policy}
\label{sec:fosa_abstract_policy}

The abstract policy must generate variable-dimensional task specifications through variable-length construction sequences, since different embodiments require different entities, constraints, goals, and arguments. FOSA parameterizes this policy as a GFlowNet and trains it with local-search sampling~\cite{kim2024local}. Every complete skeleton has exactly one goal atom, which defines its goal family $g(\mathbf{x})$; FOSA learns one $\log Z_{g}$ per family in Eq.~\eqref{eq:tb-loss}, so that within each family terminal skeletons are sampled in proportion to $R(\mathbf{x})$ rather than collapsing onto a single best task.

\textbf{Abstract State.}
An abstract state at construction step $t$ is a partial skeleton
\begin{equation}
    \mathbf{x}^{t}=[E_{\mathbf{x}^{t}},I_{\mathbf{x}^{t}},G_{\mathbf{x}^{t}},U_{\mathbf{x}^{t}},\boldsymbol{\theta}_{\mathbf{x}^{t}},\Phi_{\mathbf{x}^{t}}].
\end{equation}
FOSA evaluates two deterministic predicates over this structure. The validity predicate $\mathrm{Valid}_{\mathbf{e},\mathcal{M}}(\mathbf{x}^{t})$ checks type consistency, asset support, robot compatibility, and admissible parameter ranges. The completeness predicate $\mathrm{Complete}_{\mathbf{e},\mathcal{M}}(\mathbf{x}^{t})$ checks whether all entities, goals, success predicates, reward terms, and continuous arguments required for environment compilation are specified. A skeleton is complete when it binds a goal, an entity or environment, and a constraint in at least four bindings with all required arguments set. A maximum horizon of $T_{\max}$ edits prevents redundant or circular task descriptions. $P_F$ and $P_B$ read an LLM embedding of the serialized partial skeleton, including its arguments.

\textbf{Abstract Action.}
An abstract action $\mathbf{a}^{t}$ is a symbolic edit that maps $\mathbf{x}^{t}$ to $\mathbf{x}^{t+1}$. Invalid edits are masked before sampling from $P_F(\mathbf{x}^{t+1}\mid\mathbf{x}^{t})$. The action space contains four edit classes: \texttt{add\_atom}, which appends a typed atom (capability, entity or environment, goal, or constraint) with default arguments; \texttt{compose}, which adds a composition atom that binds existing components, once the skeleton has four bindings; \texttt{refine\_arg}, which moves one numeric argument of $\boldsymbol{\theta}_{\mathbf{x}^{t}}$ by one of five evenly spaced levels between its bounds, or sets a categorical argument to another admissible value, once the skeleton is complete; and \texttt{stop}, which terminates the trajectory. The \texttt{stop} action is enabled only when $\mathrm{Complete}_{\mathbf{e},\mathcal{M}}(\mathbf{x}^{t})=1$.

\textbf{Abstract Reward.}
The abstract reward favors skeletons on the empirical capability frontier that are structurally novel and concise. For a completed and trained skeleton $\mathbf{x}$, FOSA computes the utility
\begin{equation}
\label{eq:utility}
\mathcal{V}(\mathbf{x})=
\mathcal{F}(\mathbf{x})
+\lambda_{\mathrm{nov}}\mathcal{N}(\mathbf{x})
-\lambda_{\mathrm{cost}}\mathcal{C}(\mathbf{x}),
\end{equation}
where $\mathcal{N}(\mathbf{x})=\max\!\big(0,\;\sum_{k\in\{\mathrm{atom},\mathrm{pair},\mathrm{sig}\}} w_k\,\overline{(1+n)^{-1/2}}-d(\mathbf{x})\big)$ is a count-based novelty over the skeleton's atoms, atom pairs, and action--entity--condition signatures; for each kind $k$ the bar averages $(1+n)^{-1/2}$ over the skeleton's components of that kind, $n$ being the number of earlier skeletons sharing the component, and $d(\mathbf{x})$ is a near-duplicate penalty. $\mathcal{C}(\mathbf{x})=\min\!\big(1,\,0.1\max(0,|\mathbf{x}|-4)\big)$ charges atoms beyond the minimal four. The frontier term
\begin{equation}
\label{eq:frontier-score}
    \mathcal{F}(\mathbf{x})=
    4\mathcal{J}_{\mathbf{x}}(1-\mathcal{J}_{\mathbf{x}})\cdot
    \frac{\max(0,\mathcal{J}_{\mathbf{x}}-\mathcal{J}^{0}_{\mathbf{x}})}{\mathcal{J}_{\mathbf{x}}}
\end{equation}
is positive exactly on $\mathcal{B}_{\mathbf{e}}$ with $\mathcal{F}=0$ when $\mathcal{J}_{\mathbf{x}}=0$. The first factor is the normalized variance of the final success, maximal where the policy succeeds half of the time; the second is the share of that success acquired by training. This frontier reward follows the curriculum-learning principle of targeting tasks at intermediate difficulty~\cite{florensa2018automatic,zhang2020automatic,parker2022evolving,tio2023transferable} and is maximized by tasks that training makes learnable but does not master. The positive terminal reward for the GFlowNet is
\begin{equation}
\label{eq:abs-reward}
    R(\mathbf{x})=\epsilon+\exp\big(\beta\mathcal{V}(\mathbf{x})\big),
\end{equation}
where $\epsilon>0$ ensures positivity and $\beta$ controls reward sharpness. For fairness, every baseline method in Sec.~\ref{sec:result} is scored with the same $\mathcal{V}$.

\subsection{FOSA Task Policy}
\label{sec:fosa_robot_policy}

\textbf{Task State, Action, and Reward.}
For each compiled environment $\mathcal{M}_{\mathbf{e},\mathbf{x}}$, FOSA trains a model-free task policy
\begin{equation}
    \boldsymbol{\pi}_{\mathbf{x}}
    =\arg\max_{\boldsymbol{\pi}}
    \mathbb{E}_{\mathcal{M}_{\mathbf{e},\mathbf{x}},\boldsymbol{\pi}}
    \left[\sum_{h=0}^{H}\gamma^{h}r_{\mathbf{e},\mathbf{x}}(\mathbf{o}_{h},\mathbf{a}_{h})\right].
\end{equation}
The observation and action spaces follow the robot. The LLM compiler designs the state observed by the actor, the privileged state observed by the critic, the reward $r_{\mathbf{e},\mathbf{x}}$, and the domain randomization, which exposes the policy to a rich range of dynamics for robust sim-to-real transfer.

\textbf{Language-Conditioned Policy.}
The discovery stage produces task policies paired with natural language descriptions. FOSA distills them into a single language-conditioned student policy $\boldsymbol{\pi}_{\phi}(\boldsymbol{a}\mid\mathbf{o},\mathbf{x}_{\mathrm{Lang}})$: a transformer over the observations and the task text, where $\mathbf{x}_{\mathrm{Lang}}$ is encoded by the pretrained language backbone of MiniVLA~\cite{belkhale2024minivla}. The student is trained with data aggregation (DAgger)~\cite{pmlr-v15-ross11a} on every trained task of a method's archive: it is rolled out in the archived environments, and the corresponding teacher $\boldsymbol{\pi}_{\mathbf{x}}$ labels visited observations with corrective actions. This produces one policy interface while preserving the embodiment-grounded supervision obtained during discovery. The language backbone is frozen and only the transformer and action head are trained.

\emph{Running Example.}
For a quadruped with $\mathbf{x}^{0}=\varnothing$, the abstract policy may add a robot entity to $E_{\mathbf{x}}$, add an \texttt{inclined\_ramp} asset, compose it with a \texttt{low\_friction} constraint in $\Phi_{\mathbf{x}}$, refine $\boldsymbol{\theta}_{\mathbf{x}}$ to a $20^{\circ}$ incline, and terminate with a \texttt{track\_velocity} goal in $G_{\mathbf{x}}$. The task is compiled into a slippery-slope environment, a PPO policy is trained, and the resulting $\mathcal{J}_{\mathbf{x}}$ updates the abstract policy.


\section{Experimental Results}
\label{sec:result}

We evaluate FOSA on four heterogeneous robots in simulation and the real world: a ClearPath Jackal \texttt{vehicle}, an Inspire Dexterous \texttt{hand}, a Unitree Go2 \texttt{quadruped}, and a Booster T1 \texttt{humanoid}. 

Under a fixed discovery budget, each method trains \num{256} tasks per robot in one discovery run, and the resulting task policies are distilled into a single language-conditioned policy. All methods use \texttt{Gemma-4-31B-it} for task verbalization and environment compilation, \texttt{mujoco-playground}~\cite{ZakkaK-RSS-25} for simulation, and PPO~\cite{schulman2017proximal} for task policies. The abstract reward uses $\beta=8$, $\lambda_{\mathrm{nov}}=0.4$, $\lambda_{\mathrm{cost}}=0.05$, novelty weights $(w_{\mathrm{atom}},w_{\mathrm{pair}},w_{\mathrm{sig}})=(0.4,0.4,0.2)$, and a near-duplicate threshold of \num{0.85} atom-set Jaccard similarity; further details are in the supplementary material.

\textbf{Research Questions and Metrics.}
[\textbf{Q1}] Does FOSA discover a lexically diverse task distribution? Div-SB4 and Div-SB5 are $1-\mathrm{Self\mbox{-}BLEU}$ over generated task descriptions~\cite{wang2023robogen,zhu2018texygen}; the 4-gram variant penalizes repeated phrases and the 5-gram variant repeated task templates, so higher is more diverse. [\textbf{Q2}] Does FOSA cover a broad semantic space? Dist-Embed1 and Dist-Embed2 measure word- and phrase-level dispersion in the language-embedding space~\cite{wang2023robogen,zhu2018texygen}, and Expl-Eff is the fraction of candidates that yield executable policy-training runs. [\textbf{Q3}] Does FOSA generate tasks near each embodiment's empirical capability frontier? Frontier-Prox is the mean $\mathcal{F}(\mathbf{x})$ (Eq.~\eqref{eq:frontier-score}) over generated tasks. [\textbf{Q4}] Does the discovered distribution improve downstream generalization? We measure the success and normalized episodic reward (shaped return divided by a task-specific scale) of the distilled policy on unseen tasks.

\begin{figure}[b!]
  \centering
  \vspace*{-0.1in}
  \begin{subfigure}[b]{0.245\textwidth}
    \centering
    \includegraphics[width=1\textwidth]{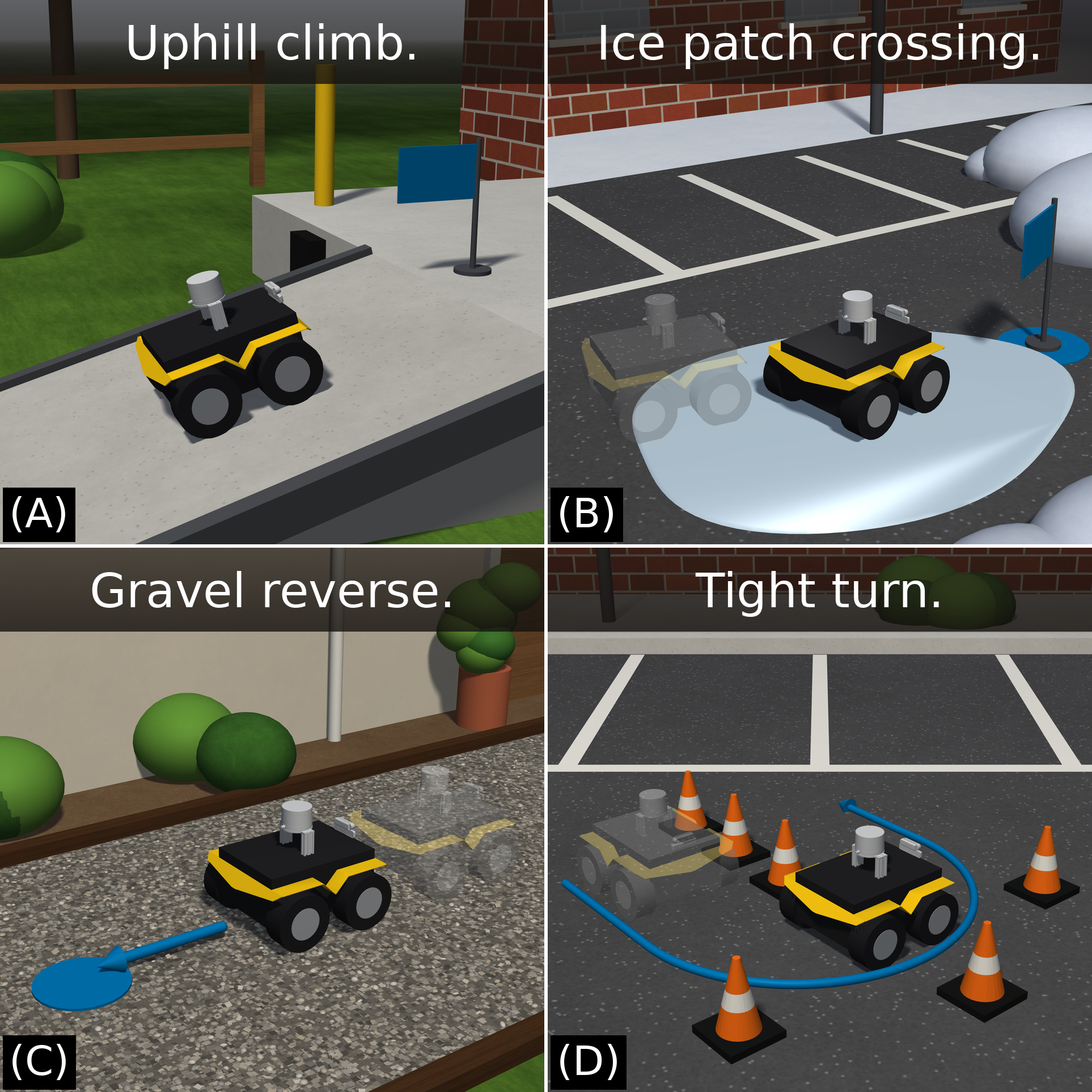}
    \vspace*{-0.225in}
    \caption*{Vehicle}
    \label{fig:ex1}
  \end{subfigure}
  \hfill
  \begin{subfigure}[b]{0.245\textwidth}
    \centering
    \includegraphics[width=1\textwidth]{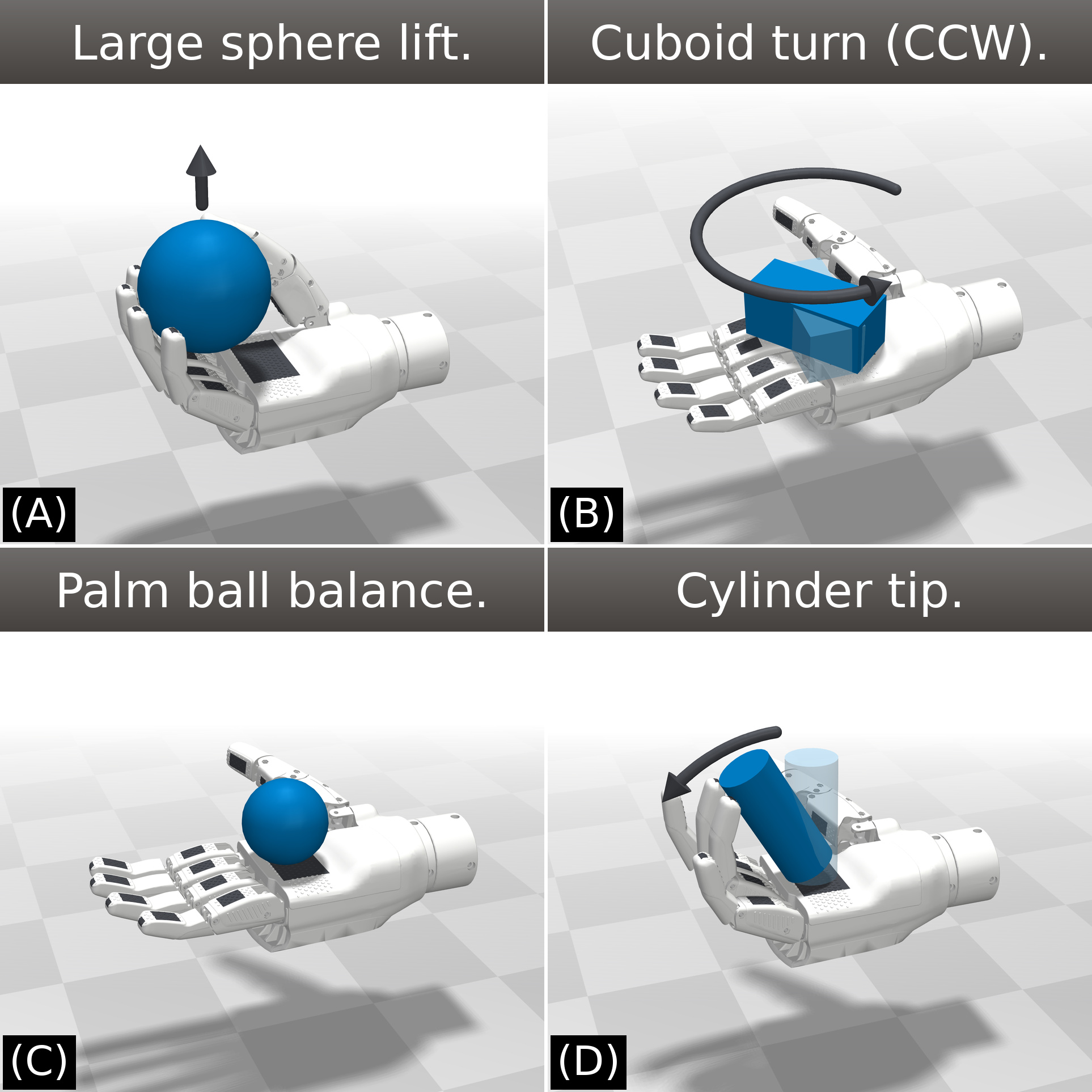}
    \vspace*{-0.225in}
    \caption*{Hand}
    \label{fig:ex2}
  \end{subfigure}
  \hfill
  \begin{subfigure}[b]{0.245\textwidth}
    \centering
    \includegraphics[width=1\textwidth]{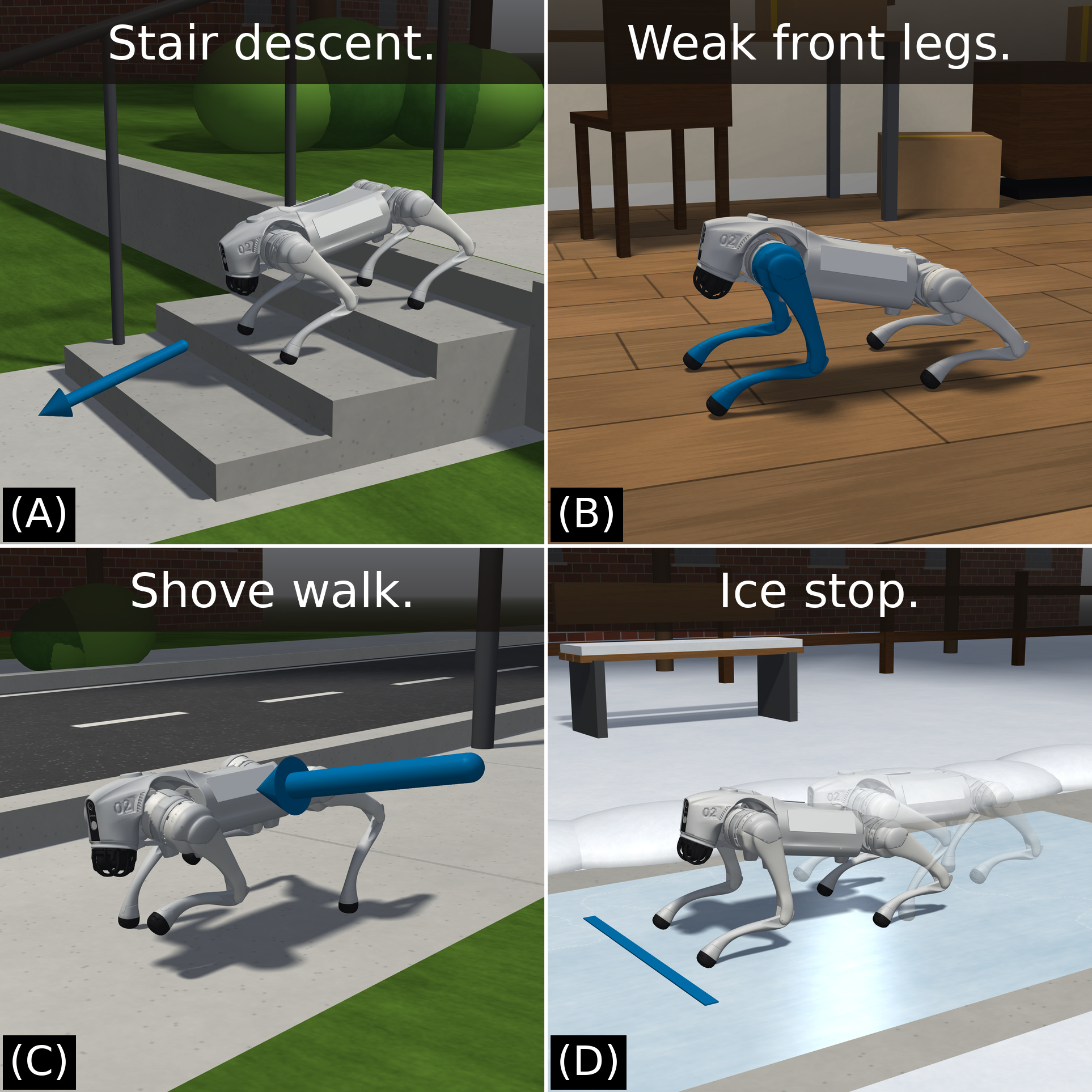} 
    \vspace*{-0.225in}
    \caption*{Quadruped}
    \label{fig:ex3}
  \end{subfigure}
  \hfill
  \begin{subfigure}[b]{0.245\textwidth}
    \centering
    \includegraphics[width=1\textwidth]{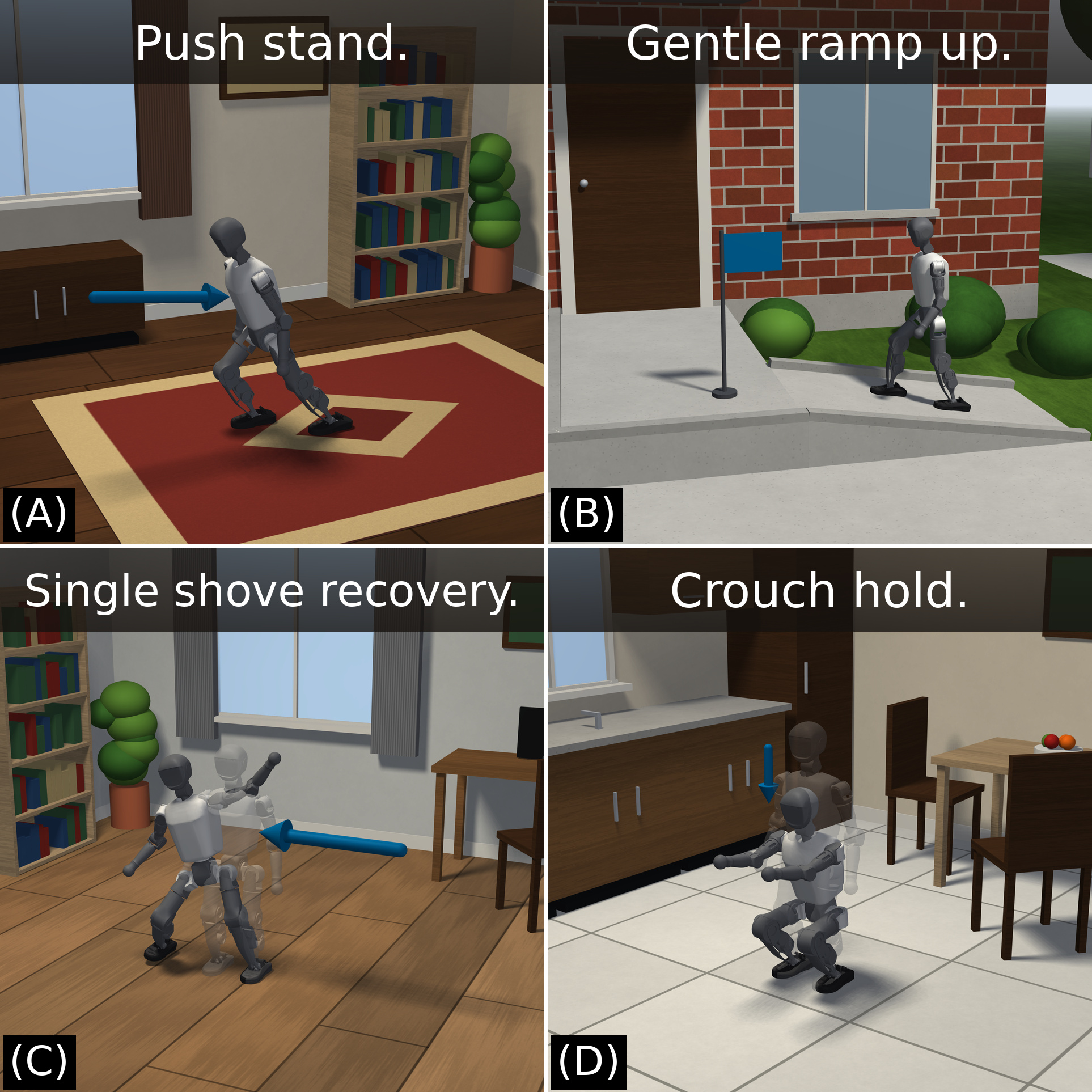}
    \vspace*{-0.225in}
    \caption*{Humanoid}
    \label{fig:ex4}
  \end{subfigure}
  \vspace*{-0.25in}
  \caption{\small Held-out evaluation tasks for the distilled language-conditioned policy across four robots.}
  \vspace*{-0.1in}
  \label{fig:sim-snapshot}
\end{figure}

\textbf{Held-Out Tasks.}
For each robot we author four held-out tasks with the LLM, disjoint from every method's discovery log (Fig.~\ref{fig:sim-snapshot}): \emph{uphill climb}, \emph{ice patch crossing}, \emph{gravel reverse}, and \emph{tight turn} (\texttt{vehicle}); \emph{large sphere lift}, \emph{cuboid turn} (counterclockwise), \emph{palm ball balance}, and \emph{cylinder tip} (\texttt{hand}); \emph{stair descent}, \emph{weak front legs}, \emph{shove walk}, and \emph{ice stop} (\texttt{quadruped}); \emph{push stand}, \emph{gentle ramp up}, \emph{single shove recovery}, and \emph{crouch hold} (\texttt{humanoid}).

\textbf{Baselines and Ablations.}
We compare against six task-discovery mechanisms. \textbf{LLM-CL} is an LLM curriculum that proposes each next task from the task history~\cite{wang2023robogen,ma2024dreureka,liang2024eurekaverse}. \textbf{LLM-MCTS} branches a Monte Carlo tree search (MCTS) with LLM-generated symbolic expansions~\cite{swiechowski2023monte}; \textbf{MCTS} removes the LLM and expands directly from the symbolic abstraction library; \textbf{MCTS-Learn} adds MuZero-style learned representation, dynamics, reward, policy, and value models~\cite{schrittwieser2020mastering}. \textbf{FOSA-RND} replaces the GFlowNet objective with random network distillation over symbolic abstractions~\cite{burda2018exploration}, and \textbf{FOSA-LLM} replaces the learned abstract policy with LLM-proposed edits. Together they isolate the effects of symbolic search, LLM priors, and the learned abstract policy.

\subsection{Simulation Results}

\begin{figure}[t]
    \centering
    \includegraphics[width=1.\linewidth]{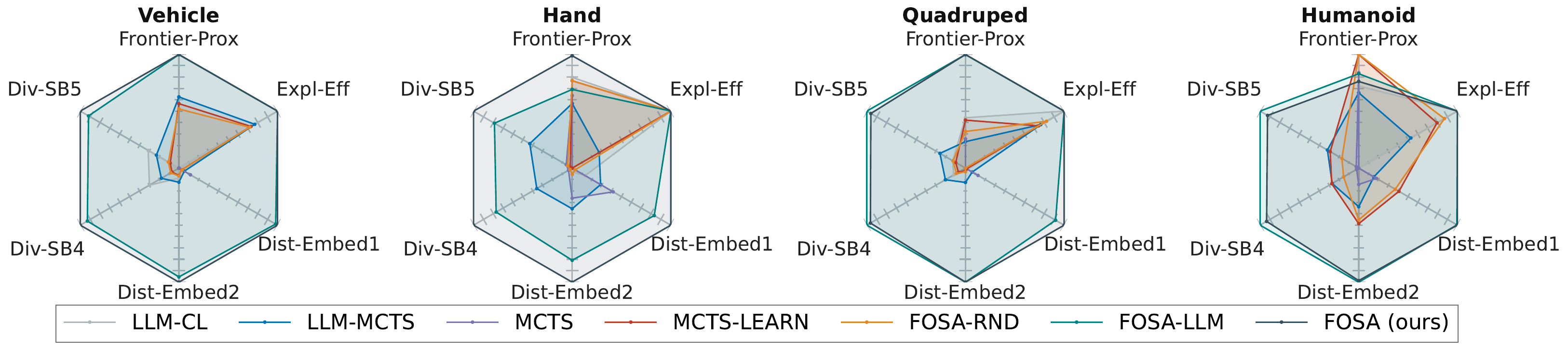}
    \vspace*{-5mm}
    \caption{\small Comparison of discovered tasks along different axes: lexical diversity (Div-SB4, Div-SB5), semantic dispersion in the language-embedding space (Dist-Embed1, Dist-Embed2), exploration efficiency (Expl-Eff), and proximity to the empirical capability frontier (Frontier-Prox).}
    \label{fig:sim-comp}
    \vspace*{-0.1in}
\end{figure}

\textbf{Can FOSA discover diverse tasks efficiently?}
Fig.~\ref{fig:sim-comp} compares discovered tasks across diversity, semantic coverage, exploration efficiency, and frontier proximity. Under the same \num{256}-task budget, FOSA has the strongest overall profile on all robots. Its gains over the MCTS variants indicate that a learned stochastic construction policy is more effective than expanding a tree with hand-specified or value-estimated branching, and its gains over the LLM-driven baselines indicate that language-level plausibility is insufficient: {\em generated tasks must also be grounded in whether the embodiment can learn them under a fixed policy-training budget}. All LLM-driven baselines use one fixed model; a stronger language prior would complement, not replace, embodiment-grounded scoring.

\textbf{Can FOSA approach empirical capability frontiers?}
FOSA discovers tasks closer to the empirical frontier on every robot except \texttt{humanoid}, where FOSA-RND scores marginally higher on Frontier-Prox. The gap reflects the larger and sparser symbolic space of whole-body humanoid control, in which an RND-style novelty bonus can occasionally reach frontier skeletons before the learned generator has accumulated reward feedback. 

\emph{Tasks that appear semantically easy can be physically hard, and vice versa.} For \texttt{hand}, a simple cube object is never reoriented even under loose orientation tolerance, whereas a ball of similar size is reoriented in more than half of the tasks. For \texttt{vehicle}, halving the wheel torque or adding more than \SI{10}{\centi\meter} of terrain relief barely changes velocity tracking success, and for \texttt{quadruped} a \SI{30}{\percent} torque loss matters only above \SI{0.5}{\meter\per\second}. Such inversions are invisible to semantic plausibility and should be identified by grounding task generation in physical policy learning.

\begin{table}[t!]
\centering
\vspace*{-0.1in}
\caption{Normalized reward of the distilled policy on held-out tasks.}
\label{tab:sim-policy}
\setlength{\tabcolsep}{3.2pt}
\renewcommand{\arraystretch}{1.18}

\newcommand{\sidecell}[2]{%
  \multirow{#2}{*}{\rotatebox[origin=c]{90}{\textit{#1}}}%
}
\newcommand{\best}[1]{{\color{blue!40!black}\mathbf{#1}}}

\begin{adjustbox}{max width=\textwidth}
\begin{tabular}{@{}c l *{16}{c}@{}}
\toprule
& \multirow{2}{*}{\textbf{Method}}
& \multicolumn{4}{c}{\textbf{Vehicle}}
& \multicolumn{4}{c}{\textbf{Hand}}
& \multicolumn{4}{c}{\textbf{Quadruped}}
& \multicolumn{4}{c}{\textbf{Humanoid}} \\
\cmidrule(lr){3-6}
\cmidrule(lr){7-10}
\cmidrule(lr){11-14}
\cmidrule(l){15-18}
&
& \textsc{A} & \textsc{B} & \textsc{C} & \textsc{D}
& \textsc{A} & \textsc{B} & \textsc{C} & \textsc{D}
& \textsc{A} & \textsc{B} & \textsc{C} & \textsc{D}
& \textsc{A} & \textsc{B} & \textsc{C} & \textsc{D} \\
\midrule

\sidecell{Baseline}{4}
& LLM-CL
& $0.64$ & $0.36$ & $0.57$ & $0.76$
& $0.29$ & $\mathbf{0.91}$ & $0.00$ & $0.57$
& $0.33$ & $0.77$ & $0.65$ & $0.28$
& $0.88$ & $0.12$ & $0.11$ & $0.11$ \\

& LLM-MCTS
& $0.48$ & $0.36$ & $0.74$ & $\best{1.00}$
& $0.61$ & $0.00$ & $0.00$ & $0.75$
& $0.51$ & $0.18$ & $\mathbf{0.97}$ & $0.18$
& $0.18$ & $0.13$ & $0.71$ & $0.06$ \\

& MCTS
& $0.62$ & $0.35$ & $\mathbf{0.86}$ & $0.85$
& $0.14$ & $\best{1.00}$ & $0.27$ & $0.61$
& $0.28$ & $0.42$ & $\mathbf{0.97}$ & $0.29$
& $0.71$ & $0.95$ & $0.93$ & $0.71$ \\

& MCTS-Learn
& $0.58$ & $\best{1.00}$ & $0.31$ & $\mathbf{1.00}$
& $0.49$ & $0.15$ & $0.00$ & $0.57$
& $0.30$ & $0.52$ & $\best{1.00}$ & $0.22$
& $\mathbf{0.88}$ & $0.97$ & $0.79$ & $\best{0.98}$ \\

\midrule

\sidecell{Ablation}{3}
& FOSA-RND
& $0.42$ & $\mathbf{0.98}$ & $0.80$ & $0.37$
& $0.59$ & $0.55$ & $0.00$ & $0.29$
& $0.28$ & $0.28$ & $\mathbf{0.99}$ & $0.16$
& $\best{1.00}$ & $0.13$ & $\best{1.00}$ & $\mathbf{1.00}$ \\

& FOSA-LLM
& $\mathbf{0.70}$ & $0.84$ & $\best{1.00}$ & $0.83$
& $\best{1.00}$ & $0.58$ & $\mathbf{0.94}$ & $0.41$
& $\mathbf{0.83}$ & $\best{1.00}$ & $0.71$ & $0.95$
& $0.11$ & $0.11$ & $0.85$ & $0.94$ \\

& FOSA (Ours)
& $\best{1.00}$ & $0.69$ & $0.57$ & $0.87$
& $0.49$ & $0.15$ & $\best{1.00}$ & $\best{1.00}$
& $\best{1.00}$ & $\mathbf{0.97}$ & $0.95$ & $\best{1.00}$
& $\mathbf{0.97}$ & $\best{1.00}$ & $\mathbf{0.87}$ & $0.82$ \\

\bottomrule
\end{tabular}
\end{adjustbox}

\vspace*{-2pt}
\begin{flushleft}
\small
\textsc{A}--\textsc{D} follow the held-out task order listed in Sec.~\ref{sec:result} for each robot. Blue bold indicates the best method and black bold indicates the second best.
\end{flushleft}
\vspace*{-0.1in}
\end{table}

\begin{figure}[h!]
  \centering
  \includegraphics[width=0.85\textwidth]{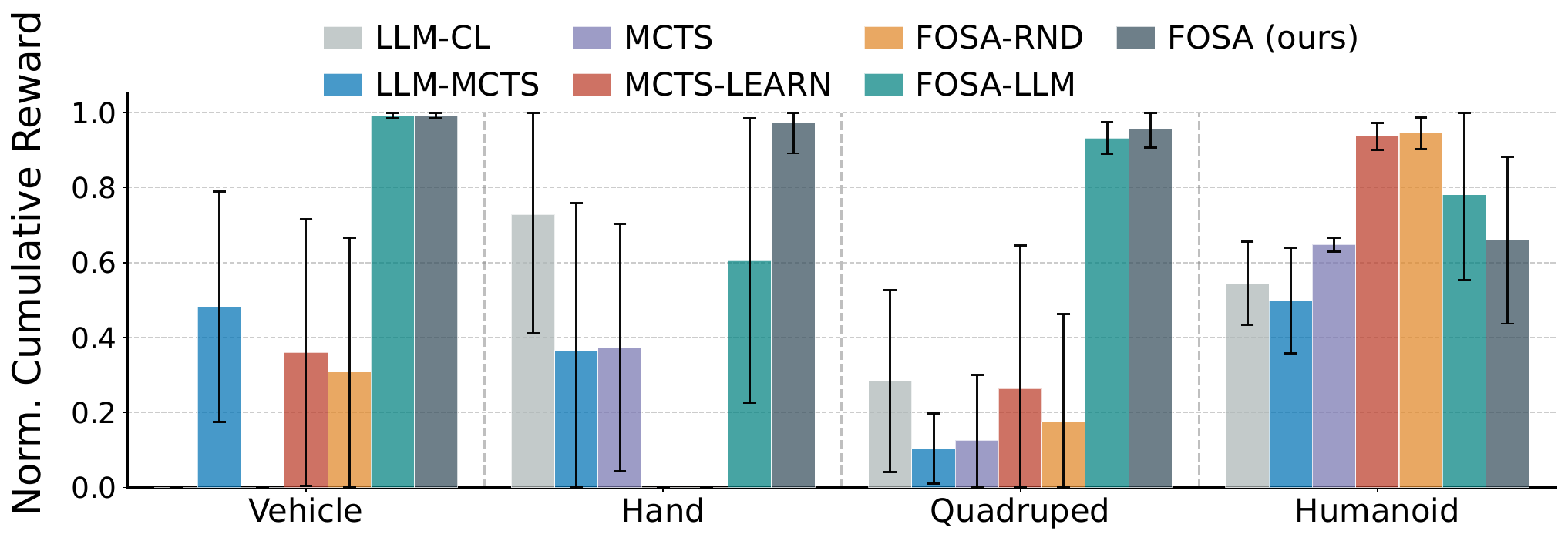}
  \vspace*{-0.1in}
  \caption{\small Simulation evaluation shows normalized cumulative reward of the distilled language-conditioned policies on the explored task suites. Each task is evaluated over \num{10} random seeds.}
  \vspace*{-0.1in}
  \label{fig:sim-policy}
\end{figure}

\textbf{Can FOSA generalize to unseen tasks?}
Fig.~\ref{fig:sim-policy} reports the distilled policy on the discovered tasks and Table~\ref{tab:sim-policy} on the \num{16} held-out tasks. Averaged over held-out tasks, the policy distilled from FOSA-discovered tasks attains the highest normalized reward (\num{0.83}), ahead of FOSA-LLM (\num{0.74}) and the strongest baseline, MCTS (\num{0.63}), with the largest margins on \texttt{quadruped} and \texttt{humanoid}, where whole-body contact makes difficulty hardest to judge from language. The ablations sharpen this: FOSA-LLM, which keeps FOSA's grammar and reward but replaces the learned $P_F$ with LLM-proposed edits, matches or exceeds FOSA on \texttt{vehicle} and \texttt{hand} yet drops to \num{0.50} on \texttt{humanoid}. A language prior thus suffices where semantic and physical difficulty coincide and fails where they diverge, which is where the reward-grounded policy matters most. Consistent with Fig.~\ref{fig:sim-comp}, methods that discover diverse tasks near the frontier also provide more useful downstream supervision.

\subsection{Algorithmic Analysis}
\label{sec:analysis}

\begin{figure}[h!]
  \centering
  \vspace*{-0.05in}
  \includegraphics[width=0.985\linewidth]{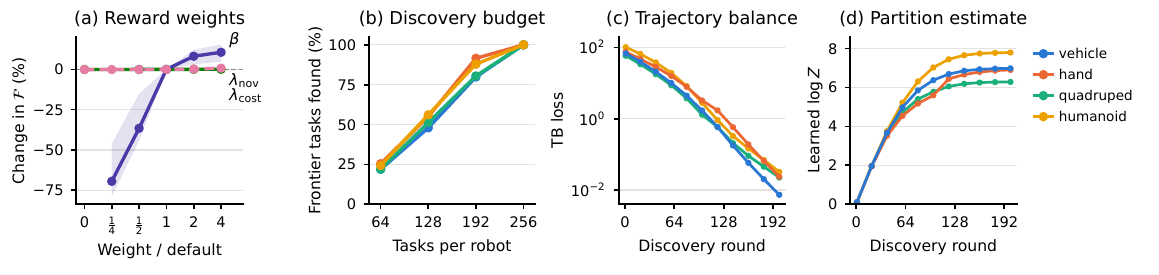}
  \vspace*{-0.15in}
  \caption{\small (a) Sensitivity to reward weights, as the change in expected frontier score of the target $P^{\star}\propto R$. (b) Sensitivity to discovery budget, as the fraction of FOSA's final frontier tasks found after its first $n$ trained tasks. (c)--(d) GFlowNet training diagnostics over discovery rounds: trajectory-balance loss (log scale) and learned $\log Z$, mean over three seeds per robot.}
  \label{fig:sensitivity}
\end{figure}

We analyze FOSA's sensitivity to its reward weights and discovery budget, and the convergence of its abstract policy (Fig.~\ref{fig:sensitivity}). Sensitivities are reported relative to FOSA's default setting.

\textbf{Reward sensitivity.}
On the distinct skeletons trained by all methods except LLM-CL, we vary one weight of $\mathcal{V}$ at a time between $\nicefrac{1}{4}\times$ and $4\times$ its default and track the expected frontier score of $P^{\star}(\mathbf{x})\propto R(\mathbf{x})$. Since $\mathcal{F}$ dominates $\mathcal{V}$, $\lambda_{\mathrm{nov}}\in[0,1.6]$ and $\lambda_{\mathrm{cost}}\in[0,0.2]$ move the score by at most \SI{2}{\percent} and \SI{0.2}{\percent}. Only $\beta$ matters: halving it lowers the score by \SIrange{15}{44}{\percent} and quartering it by \SIrange{46}{79}{\percent}, leaving \SIrange{29}{65}{\percent} of the target's mass on the frontier instead of \SIrange{98}{100}{\percent}; quadrupling it gains at most \SI{16}{\percent} but drops over a third of the goal families on \texttt{quadruped}. The default is thus the smallest $\beta$ that concentrates on the frontier without losing coverage.

\textbf{Budget sensitivity.}
Frontier tasks keep accumulating throughout discovery: after \num{64}, \num{128}, and \num{192} trained tasks, FOSA has found \SIrange{22}{25}{\percent}, \SIrange{48}{56}{\percent}, and \SIrange{80}{92}{\percent} of its final frontier tasks across robots. Frontier membership is also decided early in each task's training: rescoring $\mathcal{F}$ after \SI{56}{\percent} of the PPO budget retains a rank correlation of \numrange{0.79}{0.92} with the full-budget ranking and keeps \SIrange{85}{96}{\percent} of frontier labels, so the per-task budget could be cut by roughly \SI{40}{\percent} without changing what FOSA discovers.

\textbf{Convergence.}
On every robot, the trajectory-balance loss falls by four orders of magnitude over \num{200} discovery rounds and the learned $\log Z$ flattens after about \num{128}. On the pooled replayed skeletons, the learned flow tracks the target reward (Spearman $\rho=0.90$, Pearson $r=0.98$, MAE $0.094$ in $\log R$). These diagnostics verify the online objective empirically; the archive-dependent novelty term precludes a stationary-reward guarantee.

\subsection{Real-World Results}

\begin{figure*}[t!]
  \centering
  \begin{minipage}[c]{0.58\textwidth}
    \centering

    \makebox[0.05\linewidth][c]{\rotatebox[origin=c]{90}{\strut Vehicle}}%
    \hfill
    \begin{subfigure}[c]{0.94\linewidth}
      \centering
      \includegraphics[width=\linewidth]{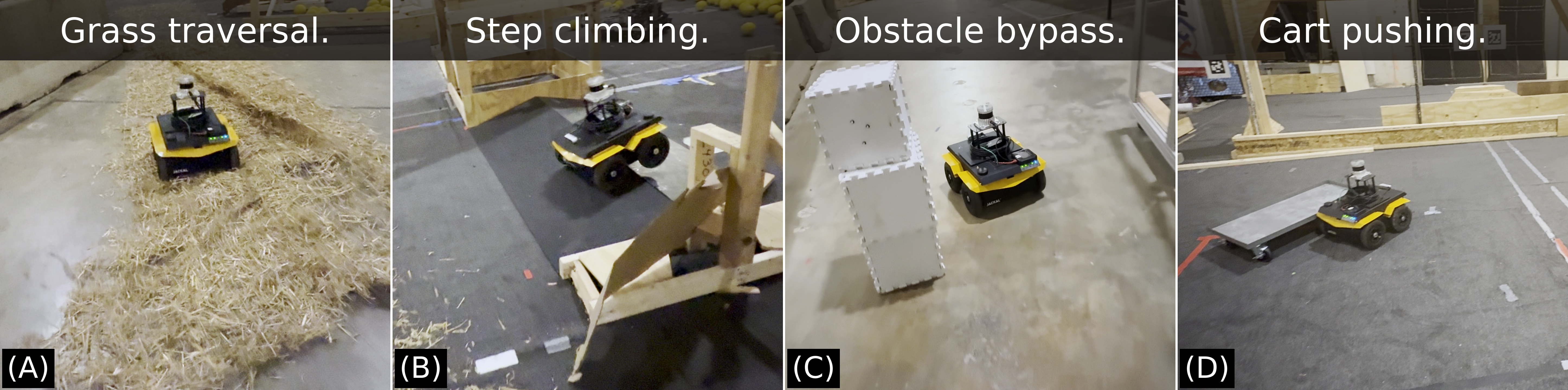}
      \label{fig:real-snapshot-vehicle}
    \end{subfigure}

    \vspace*{-11.5pt}

    \makebox[0.05\linewidth][c]{\rotatebox[origin=c]{90}{\strut Quadruped}}%
    \hfill
    \begin{subfigure}[c]{0.94\linewidth}
      \centering
      \includegraphics[width=\linewidth]{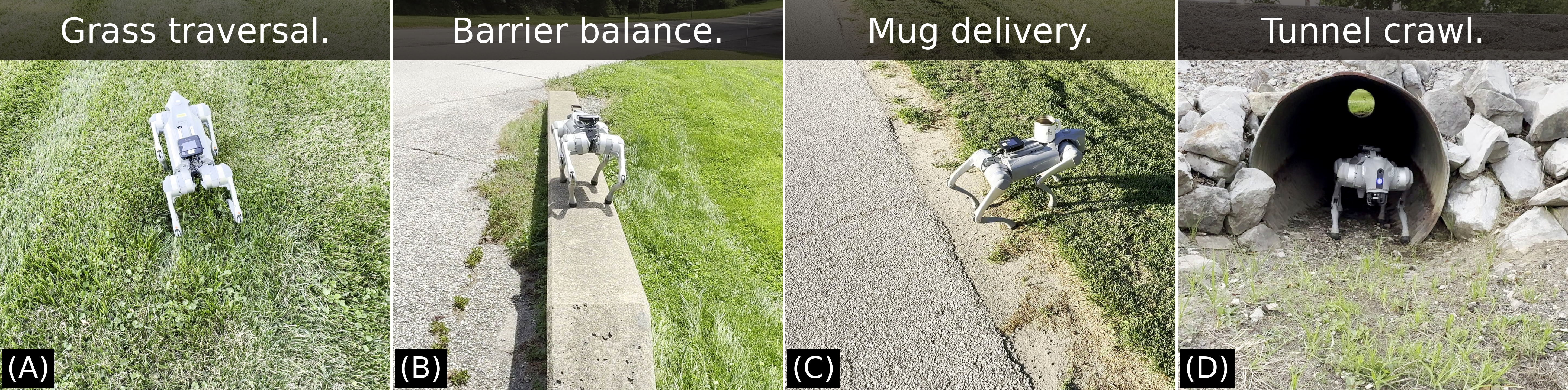}
      \label{fig:real-snapshot-quadruped}
    \end{subfigure}

    \vspace*{-0.2in}
    \caption{\small Held-out evaluation tasks for the distilled language-conditioned policy across two robots.}
    \label{fig:real-snapshot}
  \end{minipage}%
  \hfill
  \begin{minipage}[c]{0.39\textwidth}
    \centering
    \setlength{\tabcolsep}{1.8pt}
    \renewcommand{\arraystretch}{1.08}

    \newcommand{\sidecell}[2]{%
      \multirow{#2}{*}{\rotatebox[origin=c]{90}{\textit{#1}}}%
    }
    \newcommand{\best}[1]{{\color{blue!40!black}\mathbf{#1}}}

    \begin{adjustbox}{max width=\linewidth}
    \begin{tabular}{@{}c l *{8}{c}@{}}
    \toprule
    & \multirow{2}{*}{\textbf{Method}}
    & \multicolumn{4}{c}{\textbf{Vehicle}}
    & \multicolumn{4}{c}{\textbf{Quadruped}} \\
    \cmidrule(lr){3-6}
    \cmidrule(l){7-10}
    &
    & \textsc{A} & \textsc{B} & \textsc{C} & \textsc{D}
    & \textsc{A} & \textsc{B} & \textsc{C} & \textsc{D} \\
    \midrule

    \sidecell{Baseline}{4}
      & LLM-CL
      & $0$ & $\mathbf{50}$ & $0$ & $\mathbf{17}$
      & $0$ & $0$ & $0$ & $33$ \\
      
      & LLM-MCTS
      & $17$ & $33$ & $0$ & $\mathbf{17}$
      & $0$ & $0$ & $0$ & $0$ \\
      
      & MCTS
      & $0$ & $33$ & $\mathbf{17}$ & $\mathbf{17}$
      & $0$ & $0$ & $0$ & $0$ \\
      
      & MCTS-Learn
      & $17$ & $\mathbf{50}$ & $\mathbf{17}$ & $\best{33}$
      & $0$ & $0$ & $0$ & $0$ \\
      
      \midrule
      
      \sidecell{Ablation}{3}
      & FOSA-RND
      & $17$ & $\mathbf{50}$ & $0$ & $0$
      & $0$ & $0$ & $0$ & $0$ \\
      
      & FOSA-LLM
      & $0$ & $17$ & $0$ & $17$
      & $0$ & $0$ & $\mathbf{17}$ & $\mathbf{83}$ \\
      
      & FOSA (Ours)
      & $\best{50}$ & $\best{67}$ & $\mathbf{17}$ & $\best{33}$
      & $\best{17}$ & $0$ & $\best{17}$ & $\best{100}$ \\

    \bottomrule
    \end{tabular}
    \end{adjustbox}

    \vspace*{-0.05in}
    \captionof{table}{\small Real-world success rate (\%) over six trials per held-out task.}
    \label{tab:real-policy}
  \end{minipage}
\end{figure*}

We test whether capabilities discovered in simulation transfer zero-shot to novel tasks on a ClearPath Jackal and a Unitree Go2, each running on an NVIDIA Jetson Orin with an Intel RealSense D435i at \SI{30}{\Hz} for policy inference and an OS1-64 LiDAR for \SI{100}{\Hz} odometry~\cite{fasterlio,yu2023mimu}. Each robot has four held-out tasks (Fig.~\ref{fig:real-snapshot}): \emph{grass traversal}, \emph{step climbing}, \emph{obstacle bypass}, and \emph{cart pushing} for \texttt{vehicle}; \emph{grass traversal}, \emph{barrier balance}, \emph{mug delivery}, and \emph{tunnel crawl} for \texttt{quadruped}.

\textbf{Do discovered capabilities transfer to hardware?}
Table~\ref{tab:real-policy} reports success over six trials per method--task pair. The trend matches simulation: policies distilled from FOSA-discovered tasks transfer more reliably than those from baseline task sets. Absolute success is low for every method, which we attribute to the lightweight student and the restricted simulation training protocol (Sec.~\ref{sec:conclusion}). The hardest task is \emph{barrier balance}: narrow support, low-friction contact, and posture transitions amplify errors the simulator does not fully reproduce. Within these constraints, FOSA achieves the highest real-world success, supporting capability-frontier discovery as a practical source of transferable robot training.

These results support FOSA's central premise. Because the task space is combinatorial, discovery must allocate its budget to tasks that are both learnable and informative, and the two failure modes are asymmetric: a task already solved at initialization adds no behavioral coverage, whereas a task still unsolved after training yields a teacher whose actions are uninformative or inconsistent. By sampling in proportion to a reward that is positive only on the frontier, FOSA fills its archive with tasks whose teachers encode nontrivial yet reliable control, the regime in which distillation is most sample-efficient. The archive thereby contains transferable subskills (e.g., balance regulation, multi-stage navigation) that the student recombines on tasks never seen during discovery, so the gains on unseen tasks reflect better transfer of discovered capabilities as well as better discovery.


\section{Conclusion and Limitations}
\label{sec:conclusion}

We presented Generative Flow on Symbolic Abstractions (FOSA), a framework for discovering embodiment-specific capabilities by learning an abstract policy over symbolic task skeletons. FOSA compiles generated skeletons into simulation environments and uses low-level policy-learning performance as feedback to discover tasks near the robot's empirical capability frontier. Across four heterogeneous robots, FOSA improves discovery efficiency, task diversity, and frontier proximity, and the policy distilled from its tasks executes unseen instructions on real hardware.

\textbf{Limitations.}
\emph{Empirical frontier.} The frontier depends on the simulator, policy class, reward design, and training budget, although its ranking is stable across budgets (Sec.~\ref{sec:analysis}).
\emph{Sim-to-real.} The student is a lightweight language-conditioned transformer rather than a full VLA, and compute limits kept domain randomization and task training below those of dedicated sim-to-real pipelines~\cite{hwangbo2019learning,kumar2021rma}.
\emph{Safety.} FOSA targets tasks at the edge of competence, so rollouts risk falls and collisions; all hardware runs used an operator-held emergency stop and joint velocity limits, and automatic safety filtering of generated tasks is left for future work.


\clearpage


\bibliography{root}

\clearpage
\appendix
\renewcommand{\thesection}{A\arabic{section}}
\renewcommand{\thesubsection}{\thesection.\arabic{subsection}}
\renewcommand{\theequation}{\thesection.\arabic{equation}}
\renewcommand{\thefigure}{\thesection.\arabic{figure}}
\renewcommand{\thetable}{\thesection.\arabic{table}}

\numberwithin{equation}{section}
\numberwithin{figure}{section}
\numberwithin{table}{section}

\setlength{\parindent}{0pt}
\setlength{\parskip}{0.35em}

\setlist[itemize]{leftmargin=*, itemsep=0.12em, topsep=0.25em}
\setlist[enumerate]{leftmargin=*, itemsep=0.12em, topsep=0.25em}

\newcommand{\best}[1]{{\color{blue!40!black}\mathbf{#1}}}

\begin{center}
    {\LARGE\bf Generative Flow on Symbolic Abstractions for\\Embodiment Capability Discovery}\\[0.6em]
    {\Large\bf Supplementary Material}\\[1.4em]
    {\large Youwei Yu \quad Lantao Liu}\\[0.3em]
    Indiana University Bloomington\\
    \texttt{\{youwyu, lantao\}@iu.edu}
\end{center}
\vspace{1.0em}

\section{Symbolic Task Skeleton Language and Notation}
\label{sec:sup_symbolic_language}

This section expands the symbolic task representation used by FOSA. We use the same notation as the main paper: a task skeleton is
$
\mathbf{x}=[E_{\mathbf{x}}, I_{\mathbf{x}}, G_{\mathbf{x}},
U_{\mathbf{x}}, \boldsymbol{\theta}_{\mathbf{x}}, \Phi_{\mathbf{x}}]
$,
where $E_{\mathbf{x}}$ is the set of typed entities, $I_{\mathbf{x}}$ is the initial symbolic state, $G_{\mathbf{x}}$ is the goal condition, $U_{\mathbf{x}}$ is the set of referenced operator schemas, $\boldsymbol{\theta}_{\mathbf{x}}$ stores task arguments, and $\Phi_{\mathbf{x}}$ stores feasibility constraints. A partial skeleton $\mathbf{x}^{t}$ has the same structure, but some fields may be empty or only partially specified.

\textbf{Skeleton Components.}
The symbolic domain is specified by entity types $\mathcal{Y}$, predicate schemas $\mathcal{P}$, operator schemas $\mathcal{U}$, and feasibility constraints $\Phi$. Given a finite entity set $E$, the predicate schemas induce ground atoms $\mathrm{Atoms}(\mathcal{P},E)$ and literals $\mathrm{Lits}(\mathcal{P},E)$. The fields of a skeleton are summarized in Table~\ref{tab:sup_skeleton_components}.

\begin{table}[h]
\centering
\small
\vspace*{-0.10in}
\caption{Components of a symbolic task skeleton.}
\label{tab:sup_skeleton_components}
\begin{tabular}{p{0.12\linewidth} p{0.33\linewidth} p{0.45\linewidth}}
\toprule
\textbf{Field} & \textbf{Meaning} & \textbf{Examples} \\
\midrule
$E_{\mathbf{x}}$
& Typed entities in the task.
& robot, target, object, payload, ramp, obstacle, terrain patch, door, tool \\

$I_{\mathbf{x}}$
& Initial symbolic state induced by the reset distribution.
& initial robot pose, object pose, terrain condition, contact relation, visibility condition \\

$G_{\mathbf{x}}$
& Goal literals or success conditions.
& reach target, track velocity, lift object, maintain balance, rotate object, avoid obstacle \\

$U_{\mathbf{x}}$
& Operator or task-family schemas referenced by the skeleton.
& navigate, climb, push, grasp, lift, rotate, balance, transport \\

$\boldsymbol{\theta}_{\mathbf{x}}$
& Continuous or discrete arguments needed to instantiate the task.
& slope angle, friction, object mass, target distance, velocity command, tolerance, duration \\

$\Phi_{\mathbf{x}}$
& Feasibility, safety, and simulator constraints.
& valid pose ranges, contact constraints, termination thresholds, admissible friction ranges, collision constraints \\
\bottomrule
\end{tabular}
\vspace*{-0.10in}
\end{table}

A skeleton is not only a semantic description; it must also be supported by the embodiment and backend. For example, a quadruped can instantiate terrain, balance, and locomotion-related skeletons, while a dexterous hand can instantiate grasping, lifting, and in-hand manipulation skeletons.

\textbf{Partial Skeletons and Construction.}
FOSA constructs skeletons through a sequence of symbolic edits. The initial state is the empty skeleton $\mathbf{x}^{0}=\varnothing$, and a construction trajectory is
$
\boldsymbol{\tau}=(\mathbf{x}^{0}\rightarrow\mathbf{x}^{1}\rightarrow\cdots\rightarrow\mathbf{x}^{T}=\mathbf{x})
$.
Each intermediate state $\mathbf{x}^{t}$ is a partial skeleton. A partial skeleton may already contain some entities, predicates, goals, operators, arguments, or constraints, but it is not evaluated until it becomes complete.

A complete skeleton must contain enough information to define an RL environment: it must specify the relevant entities, a nonempty goal condition, any operator schemas needed to interpret the goal, all required task arguments, and the constraints needed to instantiate the reset distribution, reward, success predicate, and termination conditions. The argument vector $\boldsymbol{\theta}_{\mathbf{x}}$ contains task-specific parameters. Different task families expose different arguments. For example, locomotion tasks may use slope, roughness, and commanded velocity, while manipulation tasks may use object size, object mass, friction, and rotation angle.

\textbf{Validity and Completeness.}
The validity predicate checks whether a partial skeleton is syntactically well-formed and supported by the robot and simulator:
\begin{equation*}
    \mathrm{Valid}_{\mathbf{e},\mathcal{M}}(\mathbf{x}^{t})
    =
    \mathbb{I}
    \left[
    \begin{array}{l}
    E_{\mathbf{x}^{t}} \text{ uses supported entity types,} \\
    I_{\mathbf{x}^{t}} \subseteq \mathrm{Atoms}(\mathcal{P},E_{\mathbf{x}^{t}}), \\
    G_{\mathbf{x}^{t}} \subseteq \mathrm{Lits}(\mathcal{P},E_{\mathbf{x}^{t}}), \\
    U_{\mathbf{x}^{t}} \text{ uses supported operator schemas,} \\
    \boldsymbol{\theta}_{\mathbf{x}^{t}} \text{ lies in admissible domains,} \\
    \Phi_{\mathbf{x}^{t}} \text{ uses supported feasibility constraints.}
    \end{array}
    \right].
\end{equation*}
The completeness predicate checks whether a valid partial skeleton contains all information needed for environment compilation:
\begin{equation*}
    \mathrm{Complete}_{\mathbf{e},\mathcal{M}}(\mathbf{x}^{t})
    =
    \mathbb{I}
    \left[
    \begin{array}{l}
    \mathrm{Valid}_{\mathbf{e},\mathcal{M}}(\mathbf{x}^{t})=1, \\
    G_{\mathbf{x}^{t}}\neq \varnothing, \\
    \text{all required arguments in } \boldsymbol{\theta}_{\mathbf{x}^{t}}
    \text{ are specified,} \\
    \text{a reset distribution, reward, success predicate, and termination} \\
    \text{condition can be induced from } \mathbf{x}^{t}.
    \end{array}
    \right].
\end{equation*}

\textbf{Canonical Form and Duplicate Structure.}
Different construction orders can lead to the same skeleton. We identify a skeleton by the sorted multiset of its bindings (atom, role, and arguments), which removes construction-order artifacts. All reported metrics count a skeleton once under this identity. FOSA's sampler tracks skeletons by construction path, so local search can re-derive and re-train an earlier skeleton (Sec.~\ref{sec:fosa_abstract_policy}). Before compilation, every grammar-based method rejects a skeleton whose key or description exactly matches an archived one. Let $S(\mathbf{x})$ be the set of atoms of $\mathbf{x}$. Near-duplicates, with
\begin{equation}
\label{eq:sup_jaccard}
    \mathrm{Jac}(\mathbf{x},\mathbf{y})
    =
    \frac{|S(\mathbf{x})\cap S(\mathbf{y})|}
         {|S(\mathbf{x})\cup S(\mathbf{y})|}\geq 0.85
\end{equation}
against a recent skeleton $\mathbf{y}$ and no changed numeric argument, are not rejected but reduce the novelty term $\mathcal{N}$, so meaningful variations in difficulty remain available.

\section{Experiment Results}
\label{sec:sup_task_details}

This section first provides details for the baseline approaches and the \num{16} held-out evaluation tasks. It then shows additional results from simulation and hardware transfer.

\subsection{Baseline Implementation Details}
\label{sec:sup_baselines}

\textbf{LLM-CL.}
LLM-CL is a direct language-model curriculum baseline. At each iteration, the LLM receives the robot and backend capability context, together with previous task descriptions and policy performance history. It then proposes one concrete next task.

\begin{center}
\setlength{\fboxsep}{6pt}
\fcolorbox{black!20}{black!3}{
\begin{minipage}{0.94\linewidth}
\small
\textbf{LLM-CL prompt template.}

\textbf{System message:}
\begin{quote}
You design feasible robot reinforcement-learning curricula for the active simulator backend. Output JSON only.
\end{quote}

\textbf{User message:} a JSON object with the following fields:
\begin{itemize}
    \item \texttt{backend}, \texttt{robot}, \texttt{robot\_display\_name}, \texttt{generator\_robot\_name}: identifiers for the active robot.
    \item \texttt{robot\_profile}, \texttt{backend\_capabilities}, \texttt{simulator\_capability\_manifest}: embodiment description and the simulator contract.
    \item \texttt{registered\_task\_adapters}, \texttt{backend\_supported\_families}, \texttt{applicable\_task\_families}, \texttt{unsupported\_families\_with\_reasons}: what the backend can instantiate.
    \item \texttt{objective}: ``Propose the next task at the robot's capability frontier: one it cannot do yet (success\_rate near 0 at the start of training) but can learn within the budget (success\_rate between 0.1 and 0.9 at the end). Tasks already solved, and tasks that stay at zero, are both worthless.''
    \item \texttt{history}: the last 20 tasks, with description, status, success rate at the start and end of training, normalized reward, $\mathcal{F}$, and errors, including rejected proposals.
    \item \texttt{previous\_validation\_issue}: the reason the previous proposal was rejected, if any.
\end{itemize}
If the prompt is too long, the profile and manifest, then the capability lists, are dropped, and the history is shortened to 10 and then 5 tasks.

\textbf{Instructions:} propose exactly one next concrete robot task; use only backend-supported capabilities and the simulator manifest; select \texttt{task\_id} from the registered adapters; describe only the goal and the objects physically needed; vary physical context only when it changes the learning objective or is supported by recent outcomes; return JSON only.

\textbf{Expected output schema:} \texttt{task\_id}, \texttt{execution}, \texttt{description}, \texttt{module\_name}, \texttt{rationale}, \texttt{expected\_difficulty}, \texttt{family\_hint}.
\end{minipage}
}
\end{center}

\textbf{MCTS.}
MCTS searches the same atom grammar as FOSA without LLM-based branching. Tree nodes are partial skeletons; the root's children add one capability atom per execution primitive.
\begin{itemize}
    \item {Selection.} UCT with score $\bar{v}+1.4\sqrt{\ln(N+1)/(n+1)}+0.25\,\mathbb{I}[n=0]+0.15\cdot4\bar{p}(1-\bar{p})$, where $\bar{p}$ is the node's mean success rate, and progressive widening admits a new child while the number of children is below $2\sqrt{N+1}$.
    \item {Expansion.} Up to four children: an atom added to an open slot, or a refinement that moves one numeric argument by one of five levels or sets a categorical argument to another admissible value.
    \item {Simulation.} A complete skeleton is verbalized, compiled, and trained with the same PPO budget as every method.
    \item {Backpropagation.} The node's utility $\mathcal{V}$ (Eq.~\eqref{eq:utility}), or $-0.25$ for a run that produced nothing, is added to every node on the path.
\end{itemize}

\textbf{LLM-MCTS.} The LLM proposes eight child skeletons for the selected node, given the parent, the grammar atoms, the most valuable atoms, and recent and failed tasks. Atom names are snapped to the nearest grammar atom, and the four highest-scoring valid children are kept. The LLM's own description is used as the task text.

\textbf{MCTS-Learn.} MuZero-style representation, dynamics, reward, policy, and value networks (\num{128}-dimensional latent) are trained on the tree's statistics with a one-step unroll. They act over eleven edit-operator types, and a \num{32}-simulation latent search supplies a prior and value that are combined with the empirical node values.

\textbf{FOSA-RND and FOSA-LLM.} Both sample from $\mathbf{x}^{0}$ without training. FOSA-RND draws every edit uniformly among valid edits; FOSA-LLM adds FOSA's LLM edit prior to the uniform logits.

\subsection{Real-World Results}
\label{sec:real_res}

The real-world evaluation covers the two available hardware platforms: the ClearPath Jackal vehicle and the Unitree Go2 quadruped. The hand and humanoid experiments are evaluated in simulation only. Although recent progress in sim-to-real transfer makes hardware deployment of these policies plausible, a detailed real-world evaluation of these embodiments is left for future work.

\begin{figure}[h]
    \centering
    \includegraphics[width=0.85\linewidth]{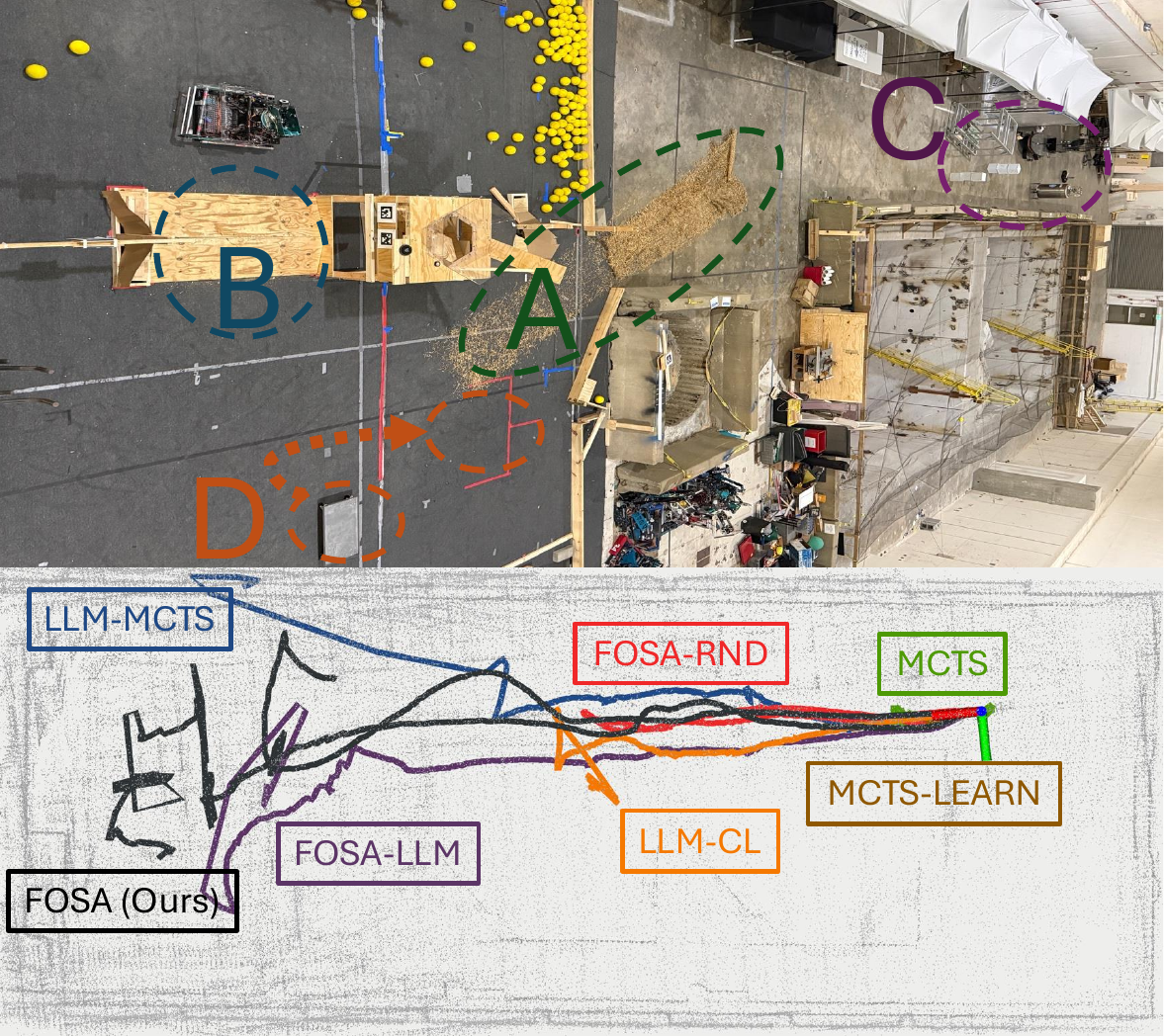}
    \vspace*{-0.08in}
    \caption{Real-world vehicle trajectories for FOSA, baselines, and ablations.}
    \label{fig:real_vehicle}
\end{figure}

\textbf{Vehicle.}
Fig.~\ref{fig:real_vehicle} shows the real-world vehicle trajectories of all compared methods in the reconstructed pointcloud map. The route consists of four stages: task C, obstacle bypass; task A, grass traversal; task B, step climbing; and task D, cart pushing. FOSA is the only method that completes the full route. The baseline methods usually fail before reaching task A, often due to obstacles or scene configurations that are not well covered during capability discovery. This trend is consistent with the main-paper results: baselines either discover less diverse task sets or fewer tasks near the empirical capability frontier.

The final cart-pushing stage is especially challenging for the ablations because the cart introduces nontrivial wheel dynamics and delayed contact effects. Although FOSA succeeds, the rollout also exposes limitations of the distilled language-conditioned policy: the vehicle sometimes produces jerky motions in the unstructured environment. Overall, these results suggest that FOSA benefits from discovering diverse, frontier-proximal capabilities, while mobile robot policies still face challenges when interacting with complex real-world objects.

\textbf{Quadruped.}
In the grass-traversal task, most methods struggle with uneven terrain and traction variation, indicating that language-level task plausibility alone does not reliably capture the physical difficulty of the scene. In the barrier-balance task, no method fully completes the task, but FOSA makes the most progress: the policy is able to maintain balance on the narrow barrier without immediately falling off.

The mug-delivery task is intuitively simple, but it requires the robot to stabilize a carried object while traversing uneven terrain. Only FOSA and its ablation FOSA-LLM keep the mug stable through the traversal; the baselines lose it at the first terrain transition. In the tunnel crawl task, all methods can pass through the tunnel, but only FOSA and its ablation lower the body height as intended. We also observe that the FOSA policy can pass the tunnel using a reverse-motion strategy, suggesting that the distilled policy can exploit non-obvious behaviors discovered during capability generation.

\textbf{Computation Time.}
Each task-policy run trains for $2\times10^{8}$ environment steps on \num{4096} parallel environments. With four runs sharing one GPU and four GPUs per discovery job, a run takes a median of \SI{1.1}{\hour} (quadruped), \SI{1.4}{\hour} (vehicle), \SI{2.1}{\hour} (hand), and \SI{2.2}{\hour} (humanoid) of wall-clock time. Distilling one student takes about \SIrange{13}{19}{\hour} on four GPUs.

\subsection{Simulation Results}
\label{sec:sim_res}

\textbf{Policy Hyperparameters.}
We use three policy-learning components: the GFlowNet abstract policy for skeleton generation, PPO task policies for grounding each compiled environment, and a MiniVLA-based language-conditioned student policy for distillation.

\begin{table}[h]
\centering
\small
\caption{Hyperparameters of the three learned components.}
\label{tab:sup_hparams}
\begin{tabular}{p{0.45\linewidth} p{0.45\linewidth}}
\toprule
\textbf{Hyperparameter} & \textbf{Value} \\
\midrule
\multicolumn{2}{l}{\textbf{FOSA Abstract Policy (GFlowNet)}} \\
\midrule
Maximum construction depth $T_{\max}$ & \num{10} \\
State encoder & \texttt{Qwen2.5-0.5B} \\
$P_F$, $P_B$ architecture & MLP $(256, 256)$ \\
Optimizer & Adam; lr $10^{-3}$ ($P_F$, $P_B$), $10^{-1}$ ($\log Z$) \\
Replay buffer & \num{1024} \\
Replay batch size & \num{32} \\
Training updates per round & \num{300} \\
Reward sharpness $\beta$ & 8.0 \\
Reward offset $\epsilon$ & $10^{-4}$ \\
Utility weights & $\lambda_{\mathrm{nov}}=0.4$, $\lambda_{\mathrm{cost}}=0.05$ \\
Near-duplicate threshold & atom Jaccard \num{0.85} \\
\midrule
\multicolumn{2}{l}{\textbf{FOSA Task Policy (PPO)}} \\
\midrule
Parallel environments & 4096 \\
Environment steps per task & $2\times10^{8}$ \\
Rollout length & 20 steps (hand: 40) \\
Batch size; mini-batches & 256; 32 \\
Learning epochs per update & 4 \\
Learning rate & $3\times10^{-4}$ \\
Discount factor $\gamma$ & 0.97 (hand reorientation: 0.99) \\
GAE parameter $\lambda$ & 0.95 \\
Clipping parameter & 0.2 \\
Value loss coefficient & 0.5 \\
Entropy coefficient & $10^{-2}$ (vehicle, humanoid: $5\times10^{-3}$) \\
Max gradient norm & 1.0 \\
Actor / critic & MLP $(512, 256, 128)$ \\
Activation & swish \\
Episode length & \num{1000} (hand reorientation: \num{600}) \\
\midrule
\multicolumn{2}{l}{\textbf{Language-conditioned Policy (MiniVLA)}} \\
\midrule
Policy & Transformer, \num{3} layers, \num{8} heads, $d=256$ \\
Inputs & last \num{10} actor observations + task text \\
Text encoder & \texttt{Qwen2.5-0.5B} \\
Action chunk & 1 \\
DAgger & \num{1024} environments \\
Loss & Smooth-$\ell_1$ imitation loss \\
Optimizer & AdamW, lr $3\times10^{-4}$, weight decay $10^{-4}$ \\
Batch size & \num{4096} per GPU (micro-batch \num{512}) \\
\bottomrule
\end{tabular}
\end{table}

\begin{figure}[h!]
    \centering
    \includegraphics[width=1\linewidth]{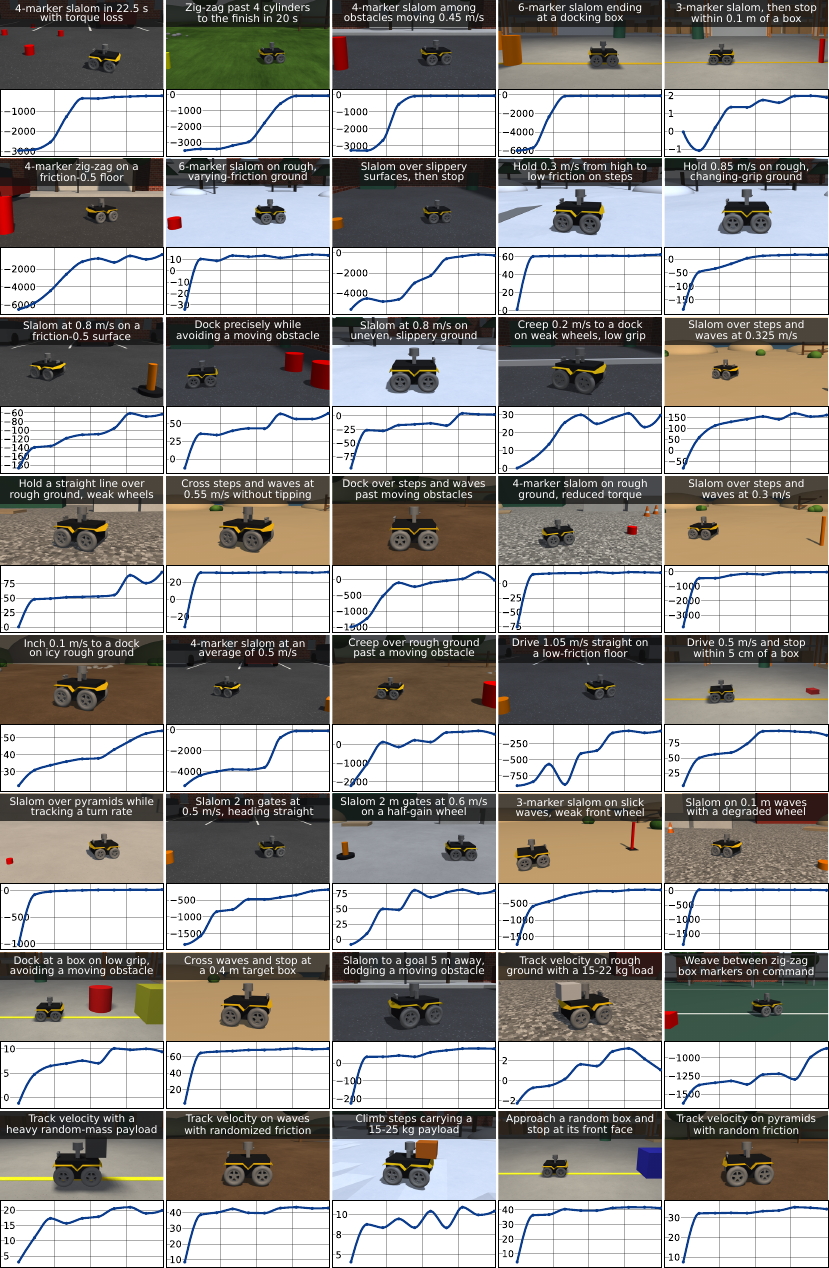}
    \vspace*{-0.15in}
    \caption{Simulation vehicle learning snapshots with task descriptions and reward curves.}
    \label{fig:frames_vehicle}
\end{figure}

\begin{figure}[h!]
    \centering
    \includegraphics[width=1\linewidth]{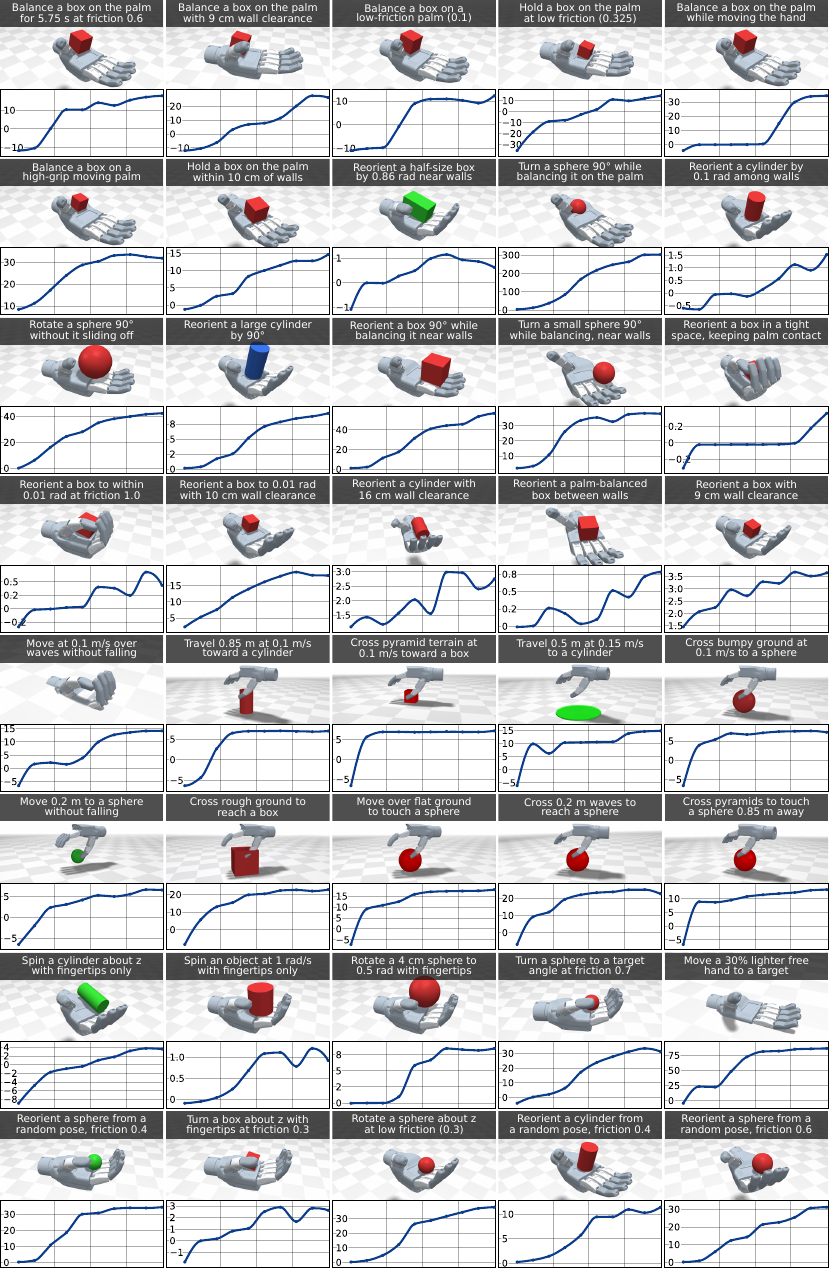}
    \vspace*{-0.15in}
    \caption{Simulation hand learning snapshots with task descriptions and reward curves.}
    \label{fig:frames_hand}
\end{figure}

\begin{figure}[h!]
    \centering
    \includegraphics[width=1\linewidth]{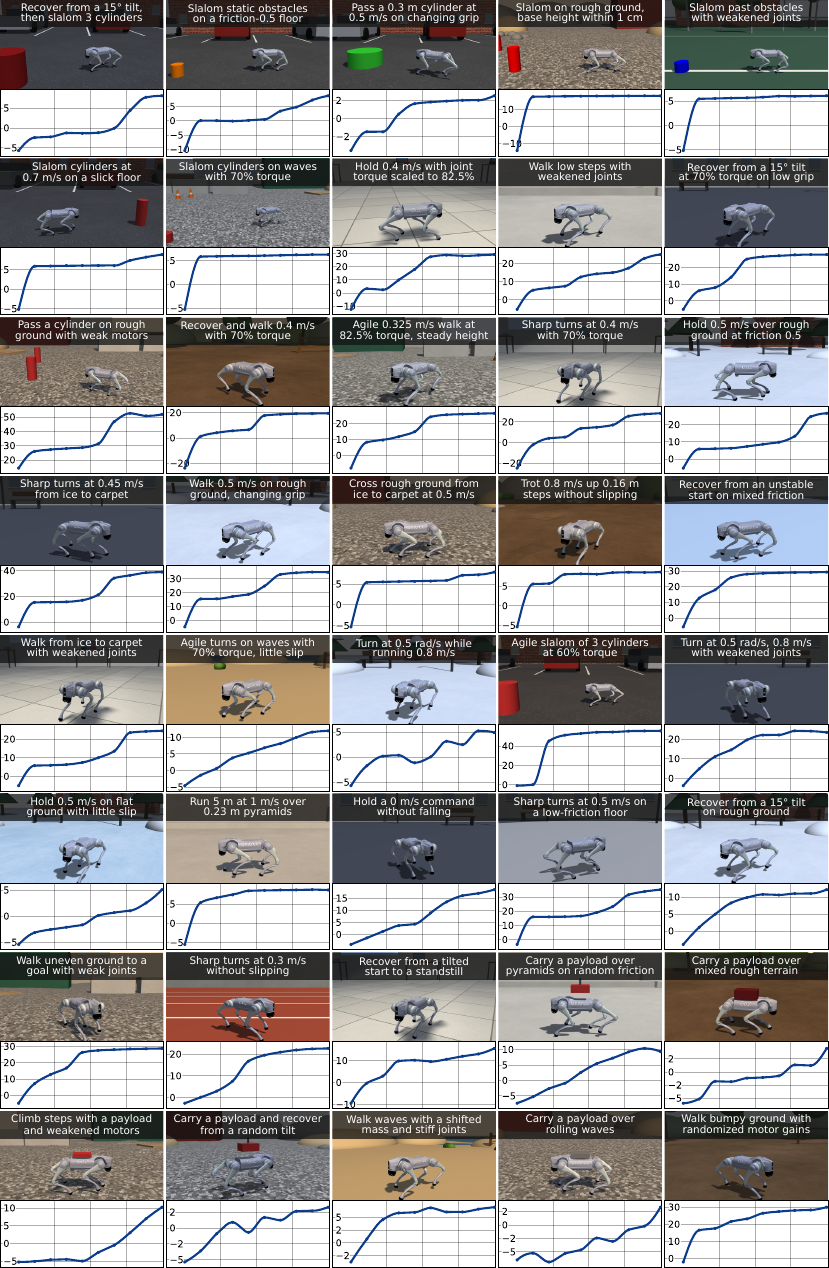}
    \vspace*{-0.15in}
    \caption{Simulation quadruped learning snapshots with task descriptions and reward curves.}
    \label{fig:frames_quadruped}
\end{figure}

\begin{figure}[h!]
    \centering
    \includegraphics[width=1\linewidth]{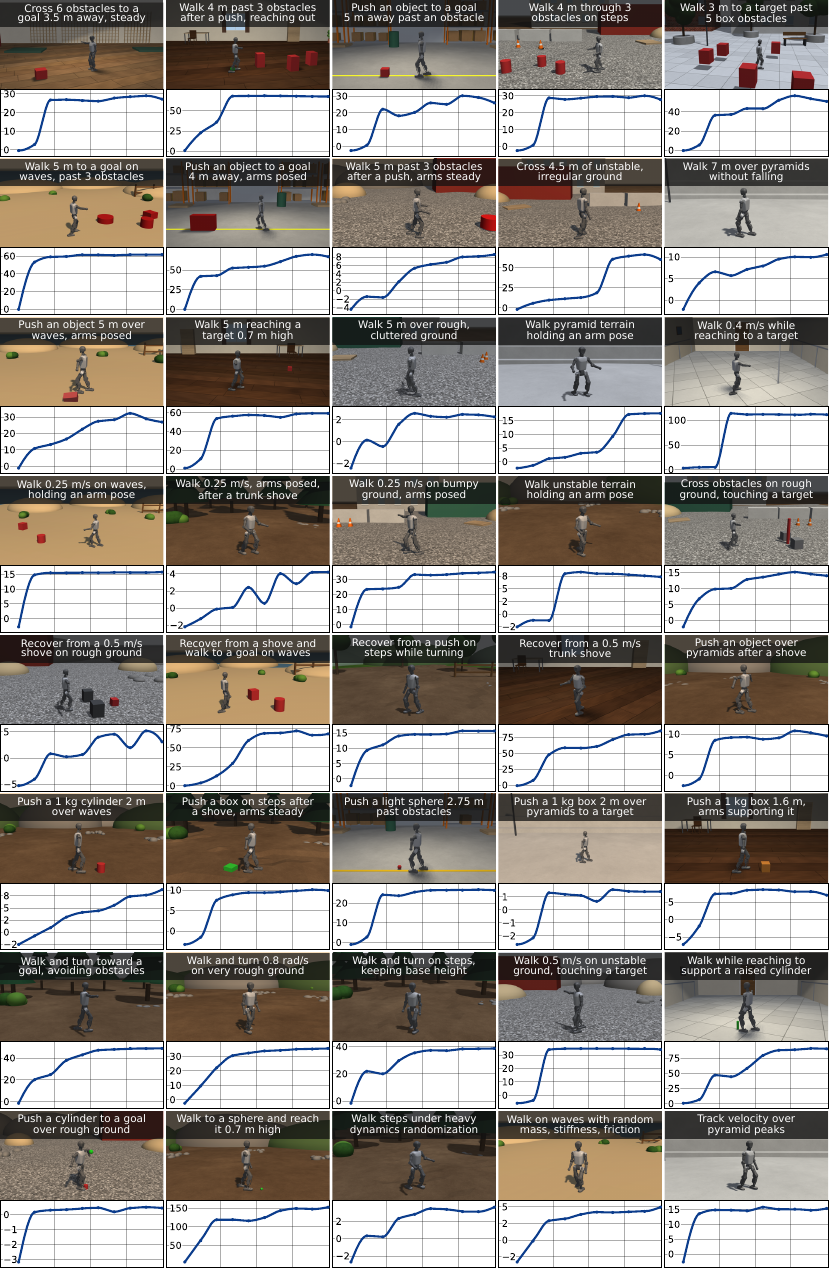}
    \vspace*{-0.15in}
    \caption{Simulation humanoid learning snapshots with task descriptions and reward curves.}
    \label{fig:frames_humanoid}
\end{figure}

\textbf{Training Logs.}
Figs.~\ref{fig:frames_vehicle}, \ref{fig:frames_hand}, \ref{fig:frames_quadruped}, and \ref{fig:frames_humanoid} show PPO training curves and simulation rollout snapshots for the four robot embodiments. These logs illustrate the task-policy learning process used to ground generated skeletons before archive construction and policy distillation.

\end{document}